\documentclass[a4paper,fleqn]{cas-sc}
\usepackage[authoryear]{natbib}
\usepackage{color, soul}
\usepackage{amsmath}
\usepackage{hhline}
\usepackage{geometry}
\usepackage{pdflscape}
\usepackage{longtable}
\usepackage{listings}
\usepackage{setspace}
\usepackage[section]{placeins}

\definecolor{backcolour}{rgb}{0.95,0.95,0.92}
\definecolor{codegreen}{rgb}{0,0.6,0}

\lstdefinestyle{myStyle}{
    backgroundcolor=\color{backcolour},   
    commentstyle=\color{codegreen},
    basicstyle=\ttfamily\footnotesize,
    breakatwhitespace=false,         
    breaklines=true,                 
    keepspaces=true,                 
    numbers=left,       
    numbersep=5pt,                  
    showspaces=false,                
    showstringspaces=false,
    showtabs=false,                  
    tabsize=2,
}

\def\tsc#1{\csdef{#1}{\textsc{\lowercase{#1}}\xspace}}
\tsc{WGM}
\tsc{QE}

\newcolumntype{P}[1]{>{\centering\arraybackslash}p{#1}}

\usepackage{verbatim}

\begin{document}
\let\WriteBookmarks\relax
\def\floatpagepagefraction{1}
\def\textpagefraction{.001}

\shorttitle{Comparison of Waymo RO Crash Data to Human Benchmarks at 7.1M Miles}    

\shortauthors{Kusano et al.}  

\title [mode = title]{
  Comparison of Waymo Rider-Only Crash Data to Human Benchmarks at 7.1 Million Miles
}

\author[1]{Kristofer D. Kusano}[
  orcid=0000-0003-4976-6114
]
\ead{kriskusano@waymo.com}
\credit{Conceptualization, Data curation, Formal analysis, Writing - original draft}
\cormark[1]

\author[1]{John M. Scanlon}[
]
\ead{johnscanlon@waymo.com}
\credit{Conceptualization, Data curation, Writing - review \& editing}

\author[1]{Yin-Hsiu Chen}[]
\ead{yinhsiuchen@waymo.com}
\credit{Methodology, Validation}

\author[1]{Timothy L. McMurry}[
  orcid=0000-0001-5912-5092
]
\ead{tmcmurry@waymo.com}
\credit{Methodology, Validation}

\author[1]{Ruoshu Chen}[]
\ead{ruoshu@waymo.com}
\credit{Validation}

\author[1]{Tilia Gode}[]
\ead{tilia@waymo.com}
\credit{Conceptualization, Supervision}

\author[1]{Trent Victor}[
  orcid=0000-0002-9550-2411
]
\ead{trentvictor@waymo.com}
\credit{Conceptualization, Supervision, Writing - review \& editing} 

\affiliation[1]{organization={Waymo, LLC.},
            addressline={1600 Ampatheater Parkway}, 
            city={Mountain View},
            postcode={94043}, 
            state={CA},
            country={USA}}

\cortext[1]{Corresponding author}

\begin{abstract}
\phantom{    }\quad\textbf{Objectives:} This paper examines the safety performance of the Waymo Driver, an SAE level 4 automated driving system (ADS) used in a rider-only (RO) ride-hailing application without a human driver, either in the vehicle or remotely.

\textbf{Methods:} ADS crash data were derived from NHTSA’s Standing General Order (SGO) reporting over 7.14 million RO miles through the end of October 2023 in Phoenix, AZ, San Francisco, CA, and Los Angeles, CA and compared to human benchmarks from the literature.

\textbf{Results:} When considering all locations together, the \textit{any-injury-reported} crashed vehicle rate was 0.6 incidents per million miles (IPMM) for the ADS vs 2.80 IPMM for the human benchmark, an 80\% reduction or a human crash rate that is 5 times higher than the ADS rate. \textit{Police-reported} crashed vehicle rates for all locations together were 2.1 IPMM for the ADS vs. 4.68 IPMM for the human benchmark, a 55\% reduction or a human crash rate that was 2.2 times higher than the ADS rate.

\textit{Police-reported} crashed vehicle rate reductions for the ADS were statistically significant when compared in San Francisco and Phoenix, as well as combined across all locations and the \textit{any-injury-reported} reductions were statistically significant in San Francisco and in all locations. The \textit{any property damage or injury} comparison had statistically significant decrease in 3 comparisons, but also non-significant results in 3 other benchmarks. When excluding ADS crashes with a delta-V less than 1 mph (a measure of sensitivity to lower reporting threshold), about half of the ADS collisions were excluded resulting in comparisons that showed a large statistically significant reduction in all comparisons except for one comparison from San Francisco.

\textbf{Conclusions:} The statistically significant reductions in \textit{police-reported} and \textit{any-injury-reported} crash rates indicate a promising positive safety impact of ADS. The direction and significance of comparisons in the \textit{any property damage or injury} outcome group are inconclusive due to difficulties in estimating a matching human benchmark. More research is needed on defining \textit{any property damage or injury} benchmarks with clear lower reporting thresholds to reduce the systematic uncertainty in the benchmark rates. Together, these crash-rate results contribute to the continuous confidence growth, together with other methodologies, in a safety case approach.

\end{abstract}

\begin{keywords}
 Automated Driving Systems\sep
 Safety Impact Analysis\sep
 Traffic Safety\sep
 Crash Data\sep
\end{keywords}

\maketitle

\section*{INTRODUCTION}\label{section:intro}

\begin{figure}
    \centering
    \includegraphics[scale=0.18]{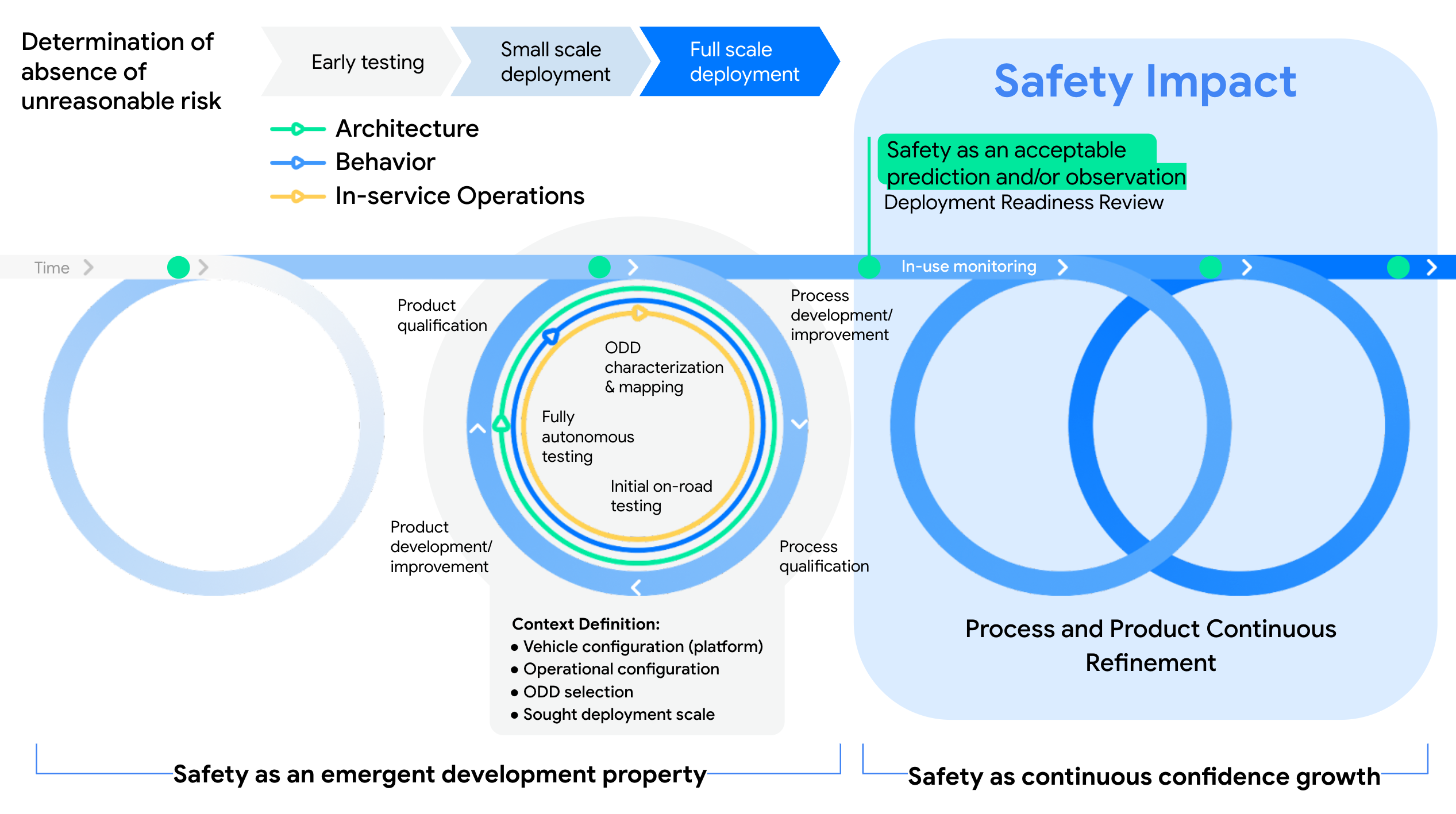}
    \caption{
      Visualization of Waymo’s Safety Determination Life Cycle
      (adapted from Figure 6, \citet{favaro2023building})}
    \label{fig:safety_lifecycle}
\end{figure}

This paper describes crash results from an in-field deployment of a Rider-Only (RO) ADS ride-hailing service as part of the safety determination lifecycle. The in-field crash results may serve as one important factor in confirming design elements and predictions done in earlier iterations of this ADS safety determination lifecycle, as shown in Figure~\ref{fig:safety_lifecycle} \citep{favaro2023building}. While a system is being developed and before starting RO operations or when considering an update to an existing RO deployment, only prospective methods which predict expected performance can be applied. In a prospective method, for example in simulated deployments \citep{webb2020readiness} or as described in \citet{favaro2023interpreting}, simulation is used to predict an ADS crash rate and compare that rate to a benchmark. In this approach, benchmarks can be set for different severity levels that are determined based on a crash severity model, as outcomes such as injury are not available from simulation. Retrospective analyses based on crash outcomes like presented in this paper can be used to complement and/or confirm these prospective methods that are used in an ADS readiness determination \citep{webb2020readiness}.

Retrospective examination of crash performance against a human benchmark contributes to the continuous confidence growth portion of an ADS safety case. Retrospective studies cannot be used to make safety determinations prior to deployment. Therefore, an ADS with a crash involvement rate lower than the general driving population does not, by itself, allow for a determination that the ADS has an absence of unreasonable risk. Assessment methodologies that use other benchmarks are important in addition to retrospective crash involvement \citep{webb2020readiness}. 
The broader safety case approach considers a range of potential issues that can arise during ADS ride-hailing operations, including measures of the ADS’s drivership, the realization of expected driving behaviors that position the ADS as a good citizen of the road. Evaluation of drivership can extend beyond crash risk.

The data sources most used to understand traffic safety historically have been derived from police-reported crash databases, like those maintained by the National Highway Traffic Safety Administration (NHTSA) and states. Recently, ADS manufacturers have been required to report crashes, albeit with a different reporting standard than that used in police-reported crash databases. 
Since July 2021, the NHTSA Standing General Order (SGO) 2021-01 has required ADS developers to report all crashes meeting certain criteria, which are then published publicly \citep{nhtsa2021sgo}. The reporting requirements for ADS manufacturers is to report any crashes that the manufacturer knows to have occurred or that are alleged to have occurred and that result in ``any property damage, injury, or fatality'', occur on a public road, and in which the ADS was engaged at any time during the 30 seconds immediately prior to the crash through the conclusion of the crash \citep{nhtsa2021sgo}.  

A retrospective safety impact assessment requires calibration of benchmark human and ADS crash and vehicle miles traveled (VMT) data sources to make a valid comparison. Given public accessibility of both human and ADS crash and VMT data, researchers have the opportunity to begin evaluating ADS safety performance. Working with these public crash and VMT data sources, however, has certain challenges. Several reviews of the literature have found that many of the early studies comparing ADS and human benchmarks introduced biases \citep{young2021critical, scanlon2023benchmark}.

\begin{figure}
    \centering
    \includegraphics[width=2.5in, keepaspectratio]{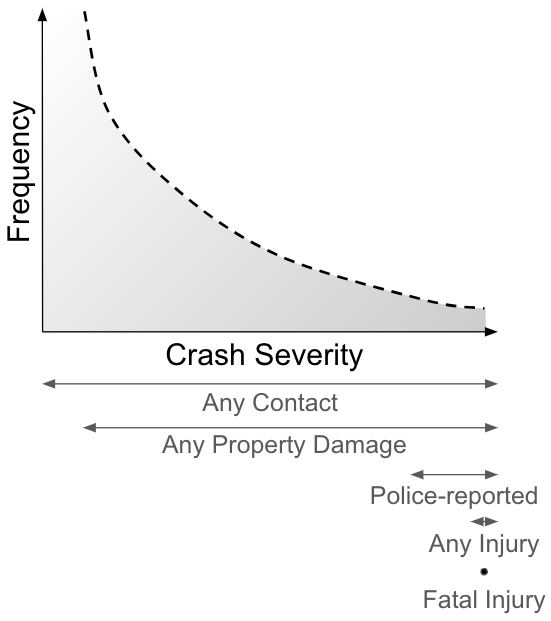}
    \caption{
      Conceptual Relationship Between Different Crash Severity Reporting Thresholds.
      The arrows below the graph show the hypothetical range of severities included in each outcome group.}
    \label{fig:reporting_thresholds}
\end{figure}

One of the most important considerations when matching benchmark and ADS crash and VMT data sources is the inclusion criteria of the data sources being compared (i.e., to correct for surveillance bias). Figure~\ref{fig:reporting_thresholds} shows the conceptual relationship between different crash severity reporting thresholds relative to their frequency of occurrence. The distribution shows hypothetical frequency by crash severity (e.g., delta-V or impact speed). Below the frequency distribution, there are ranges of severity that correspond to different reporting thresholds for crash data sources. 
The NHTSA SGO data is of the ``any property damage'' reporting threshold. Most human crash databases such as NHTSA's Crash Report Sampling Systems, CRSS, database and state crash databases are derived from crashes reported to police and/or through insurance claims. A subset of police-reported crashes can result in a reported injury (i.e., minor to fatal injuries), and a smaller subset of those any injury reported crashes result in fatal injuries. Given these different reporting thresholds, it is not appropriate to compare ADS crash data like the NHTSA SGO, which includes outcomes at the any property damage level, directly to human police-reported crash databases like national or state crash databases.

In addition to reporting thresholds, another form of bias comes from reporting bias due to underreporting. Many human crashes are not reported to police or are reported but no police report is filed. In the U.S.\, 60\% of property-damage-only and 32\% of injury crashes are not reported to police \citep{blincoe2022economic}. In contrast, ADS fleet operators are required to report almost every crash and the large number of sensors combined with operational procedures result in almost all contacts being discovered and reported. Finally, the driving environment and/or vehicle characteristics of a data source can impact the crash rate (selection bias). Factors such as population density (rural vs. urban), roadway type (surface street vs freeway), and vehicle use type (commercial heavy vehicles, passenger vehicles, ride hail vehicle) can all affect crash rates. 

Due to these potential challenges that introduce bias into comparisons of benchmark and ADS crash data, different approaches are needed to form valid comparisons depending on the nature of the human benchmark and ADS data sources. For example, if a benchmark that is derived from crashes reported to police is compared to ADS data that includes any property damage, there is a need to either apply some underreporting adjustment or restrict the ADS data to only include police-reported crashes. For ease of writing, we will use ``underreporting'' to refer to both crashes that meet the reporting threshold but were not reported and crashes that do not meet the reporting threshold. See the discussion in \citet{scanlon2023benchmark} for a more thorough discussion of these topics. An underreporting adjustment to police-reported crashes is the approach \citet{scanlon2023benchmark} took to estimate crash rates for any property damage. Alternatively, another human data source used for generating a benchmark is Naturalistic Driving Study (NDS) databases, which equip vehicles with sensors such as accelerometers and cameras to record driving. Using a combination of algorithmic triggers (e.g., hard braking or swerving) plus manual review to find crashes based on the video data, an observed crash rate can be found, such as reported in \citet{blanco2016automated} and \citet{flannagan2023establishing}.

Generating a valid comparison of human and ADS crash rates that adjusts for the above mentioned biases is of great importance at this point in time. Larger ADS deployments in recent years have led to inquiries from the general public, public officials, and researchers about the safety of current ADS fleets. All stakeholders have an interest in understanding the current state of ADS safety performance through as many objective measures as are available, which include retrospective crash rates. If biases are present in the data that either inflate or deflate the ADS or benchmark crash rates, the conclusions may lead to incorrect public perceptions of ADS safety and questionable policy recommendations or decisions. As pointed out by \citet[Appendix A.1.2]{scanlon2023benchmark} and \citet{young2021critical}, past studies have varying degrees of biases in their analyses. In addition, non-academic studies, either in the popular press or by government agencies, are being done. All data sources, both from human and ADS sources, have limitations. This paper aims to present an analysis that uses publicly available ADS and benchmark data in order to advance the understanding of ADS safety so that stakeholders can make informed decisions. This analysis is also a step toward further harmonization and standardization of retrospective safety impact analysis for ADS fleets.

In the field of traffic safety, the prevention of severe crashes is generally prioritized. One of the paradoxes of the study of crashes, however, is that severe outcomes are less frequent than low severity outcomes. This is true for several types of outcomes. For example, crashes with low dollar values of property damage outnumber high dollar values and minor injuries outnumber serious injuries. Given that ADS vehicles drive fewer miles relative to human-driven vehicles, the ADS crash data are overwhelmingly property damage only crashes. As discussed above, it is difficult to construct comparable benchmarks for any property damage crashes due to differences in reporting thresholds and underreporting. Because of the great public interest, this paper will attempt to construct a benchmark and compare it to ADS crashes for an \textit{any property damage or injury} outcome despite the challenges. Because the most-studied human crash data is either at the \textit{police-reported} or \textit{any-injury-reported} outcome level (or greater), these data sources are more well understood and are likely to have less systematic biases present compared to \textit{any property damage or injury} benchmarks. The drawback of \textit{police-reported} and \textit{any-injury-reported} outcome levels is that there will be fewer ADS crashes observed, leading to potentially larger confidence intervals. Because more serious outcome crashes often have different characteristics than less severe outcome crashes, it is possible that the safety impact and/or the confidence in the impact of an ADS may be different for different outcome levels. See the Appendix~\ref{apdx:serious_and_fatal} for further discussion on considerations for safety impact assessment for serious injury and fatal crashes.

The purpose of this paper is to present Waymo ADS crash rates from data reported according to NHTSA’s SGO and compare to benchmark rates from comparable human benchmark sources published in the literature. Comparisons were made for crashes involving \textit{any property damage or injury}, \textit{police-reported}, and \textit{any-injury-reported} outcomes. As the SGO data is publicly available, the analyses provided herein can be readily reproduced and developed by other researchers.

\section*{METHODS}\label{section:methods}

\subsection*{Matching of Benchmark and ADS Data}

Due to the data collection differences between human and ADS data, there is no set of data that represent the perfect comparison between human and ADS crash rates. Therefore, this paper aimed to develop several comparable sets of human and ADS crash rates using data published from the literature and NHTSA SGO data.

The Waymo RO ride-hailing service in Phoenix, San Francisco, and Los Angeles operates in fixed geographic areas in these cities operating 24 hours a day and 7 days a week. The ODD includes non-limited access roads with speed limits up to 50 mph and parking lots without restrictions on maneuvers. The ODD does not include severe weather conditions, such as thick fog, heavy rain, or blowing sand, but does include light rain or light fog. The ODD for the Waymo RO service has changed slightly over the years as the service expanded. The ODD features have remained largely consistent since 2022, which also corresponds to the majority of the RO miles accumulated during the study period (through October 2023). Therefore, the Waymo RO crashes described in this section were compared to crashes that occurred in the same geographical areas. The selection of human benchmarks is discussed in detail in the following section.

As NHTSA SGO ADS crashes are a mandated and publicly available data source, Waymo’s NHTSA SGO reportable crashes became the basis for the incidents examined in this study. The SGO reporting is required in all states, and thus is the most complete crash reporting required for an ADS fleet operator. Manufacturers do not report miles driven as part of the SGO data. Therefore, this study analyzed only Waymo RO crashes as these miles were the only ones available to the authors. All crashes involving Waymo operating in RO through the end of October 2023 were included. SGO data was downloaded from NHTSA's website in December 2023.

Table~\ref{tab:comparable_benchmarks} lists the ADS data and human benchmarks compared in this study and Table~\ref{tab:human_benchmarks} lists the benchmark crash rates. Comparisons were grouped into three outcome groups: \textit{Any Property Damage or Injury}, \textit{Police-reported}, and \textit{Any-Injury-Reported} outcomes. Details of each outcome group are discussed below.

\begin{table*}[width=0.95\textwidth, cols=3]
  \caption{Description of Comparable ADS and Human Benchmarks Data Sources.}
  \label{tab:comparable_benchmarks}
  \begin{tabular}{|p{4.5cm}|p{5.0cm}|p{5.0cm}|}
    \hline
    \textbf{Outcome Group} &\textbf{Waymo ADS Data} &\textbf{Human Benchmark Source} \\
    \hline
    \multirow{4}{*}{\textit{Any Property Damage or Injury}} &  All NHTSA SGO Crash - In-Transport Vehicles & \citet{scanlon2023benchmark} Any Property Damage or Injury Blincoe-Adjusted\\
    \cline{2-3}
    & NHTSA SGO Crashes - In-Transport Vehicles Excluding Low Delta-V (< 1 mph) & \citet{scanlon2023benchmark} Any Property Damage or Injury Blincoe-Adjusted\\
    \cline{2-3}
    & All NHTSA SGO Crashes in San Francisco & \citet{flannagan2023establishing}\\
    \cline{2-3}
    & All NHTSA SGO Crash in all Locations & \citet{blanco2016automated}\\
    \hline
    \textit{Police-Reported} & All NHTSA SGO Crash - In-Transport Vehicles with Police Report
    & \citet{scanlon2023benchmark} Police-Reported Unadjusted \\
    \hline
    \multirow{2}{*}{\textit{Any-Injury-Reported}} & All NHTSA SGO Crash - In-Transport Vehicles with Reported Injury
    & \citet{scanlon2023benchmark} \textit{Any-Injury-Reported} Blincoe-Adjusted \\
    \cline{2-3}
    & All NHTSA SGO Crash - In-Transport Vehicles with Reported Injury
    & \citet{scanlon2023benchmark} \textit{Any-Injury-Reported} Observed \\
    \hline
  \end{tabular}
\end{table*}

\begin{table*}[width=0.95\textwidth, cols=6]
  \caption{Crashed Vehicle Rates and VMT for Human Benchmarks.}
  \label{tab:human_benchmarks}
  \begin{tabular*}{\tblwidth}{|p{0.8in}|p{1.95in}|P{0.5in}|P{0.5in}|P{0.5in}|P{0.5in}|}
  \cline{3-6}
  \multicolumn{2}{c}{} & \multicolumn{4}{|c|}{\parbox{2in}{\centering\textbf{Human Benchmark IPMM and VMT (Millions)}}} \\
  \cline{1-6}
  \textbf{Outcome Group} & \textbf{Human Benchmark Source} & \textbf{National} & \textbf{Phoenix} & \textbf{San Francisco} & \textbf{Los Angeles} \\
  \cline{1-6}
  \multirow{6}{*}{\parbox{0.85in}{\raggedright\textit{Any Property Damage or Injury}}} & \multirow{2}{*}{\parbox[t][][t]{2in}{\citet{scanlon2023benchmark} Blincoe-adjusted \textit{Any Property Damage or Injury}}} & 9.40 & 9.43 & 10.49 & 6.84 \\
  & & 2,140,140 & 24,865 & 862 & 29,952 \\
  \cline{2-6}
  \cline{2-6}
  & \multirow{2}{*}{\citet{flannagan2023establishing} Ride-hailing NDS} & \multirow{2}{*}{\color{gray}N/A} & \multirow{2}{*}{\color{gray}N/A} & 64.9 & \multirow{2}{*}{\color{gray}N/A} \\
  & & & & 5.612 & \\
  \cline{2-6}
  & \multirow{2}{*}{\citet{blanco2016automated} SHRP-2 NDS} & 20.2 & \multirow{2}{*}{\color{gray}N/A} & \multirow{2}{*}{\color{gray}N/A} & \multirow{2}{*}{\color{gray}N/A} \\
  & & 34 & & & \\
  \cline{1-6}
  \multirow{2}{*}{\parbox{0.85in}{\textit{Police-Reported}}} & \multirow{2}{*}{\parbox[t][][t]{2in}{\citet{scanlon2023benchmark} Observed \textit{Police-Reported}}} & 4.10 & 4.31 & 5.86 & 3.39 \\
  & & 2,140,140 & 24,865 & 862 & 29,952 \\
  \cline{1-6}
  \multirow{4}{*}{\parbox{0.85in}{\textit{Any-Injury-Reported}}} & \multirow{2}{*}{\parbox[t][][t]{2in}{\citet{scanlon2023benchmark} Observed \textit{Any-Injury-Reported}}} & 1.21 & 1.24 & 3.98 & 1.54 \\
  & & 2,140,140 & 24,865 & 862 & 29,952 \\
  \cline{2-6}
  & \multirow{2}{*}{\parbox[t][][t]{2in}{\citet{scanlon2023benchmark} Blincoe-adjusted \textit{Any-Injury-Reported}}} & 1.76 & 1.81 & 5.82 & 2.26 \\
  & & 2,140,140 & 24,865 & 862 & 29,952 \\
  \cline{1-6}
  \end{tabular*}
\end{table*}

This study used two types of data sources to match \textit{Any Property Damage or Injury} benchmarks to the ADS data. First, police-reported crash data from national (the NHTSA CRSS and Fatality Analysis Reporting System, FARS) and state crash databases were adjusted for human underreporting in \citet{scanlon2023benchmark} using underreporting adjustments for property damage only and injury crashes from \citet{blincoe2022economic}. The \citet{blincoe2022economic} underreporting adjustments were developed using a combination of national police-report data, insurance claims data, and a nationally representative phone survey. These national underreporting estimates for property damage only and injury crashes were then applied to national and state crash data from 2022 in \citet{scanlon2023benchmark}. The state data was restricted to the counties where ADS deployments are currently occurring (San Francisco, CA, Maricopa, Arizona, and Los Angeles, CA). Both the state and national data were also restricted to crashes involving passenger vehicles and occurring on surface streets to match the operating conditions of the ADS. The human VMT data in \citet{scanlon2023benchmark} was sourced from Federal Highway Administration (FHWA) for national and state data using the same conditions (passenger vehicles on surface street roads). Second, observed crash rates from NDS were used. \citet{flannagan2023establishing} derived crash rates from a fleet of vehicles mostly used for ride-hailing on surface streets in San Francisco; these rates were then compared to ADS crashes in San Francisco. \citet{blanco2016automated} derived an crash rate using the SHRP-2 NDS, which were personally-owned passenger vehicles that were operated on all road types.

For the comparison of Waymo's RO crash rate to available human benchmark crash rates, the following subsets of Waymo's SGO-reported crashes were identified: (1) Those crashes reported by Waymo and without an autonomous specialist behind the wheel (RO crashes), (2) in-transport vehicles (defined as crashes where the Waymo vehicles was not parked in a parking spot), and (3) those where the Waymo vehicle was not impacted (defined as those crashes where the Waymo vehicle struck or was struck by another road user or object). In those ADS crashes with an autonomous specialist seated in the driver's seat, it is difficult to disambiguate the contributions of the ADS and the autonomous specialist. Therefore, this analysis excluded those crashes with an autonomous specialist present behind the wheel, as RO crashes are most representative of the intended autonomous ride-hailing service. The crashed vehicle count from the police reported benchmark data included only in-transport, impacted vehicles. Therefore, the \textit{any property damage or injury} benchmarks from \citet{scanlon2023benchmark} were compared with all SGO-reported ADS crashes, except those where the ADS vehicle was not-in-transport or the ADS vehicle was not impacted. Because the NDS data sources described in \citet{flannagan2023establishing} and \citet{blanco2016automated} did not specify whether not in-transport NDS vehicles were excluded, we compared these NDS data to all reported SGO ADS crashes excluding when the ADS vehicle was not impacted. All SGO collisions that met the conditions were included in the study, regardless of crash mode or contacted object.

In addition to examining all NHTSA SGO-reported crashes, results were also reported for a dataset that excluded minor crashes, which we will refer to as ``SGO Crash Excluding Low Delta-V.'' Although the benchmark data was either adjusted for underreporting or used NDS data designed to find most contacts, there is still uncertainty about the lower reporting threshold in the benchmark data and underreporting corrections. This result of excluding low delta-V ADS crashes was presented alongside the larger all SGO-reported crashes group to investigate the sensitivity of lower reporting threshold on the comparison to the benchmark. We defined these minor contacts as those in which a crash reconstruction model found that both vehicles had a change in velocity (i.e., a delta-V) of less than 1.0 mph. Crash delta-V was determined by a crash model (for vehicle-to-vehicle crashes) or estimated by collision reconstruction techniques (for fixed object crashes).

The human crash benchmarks reported in \citet{scanlon2023benchmark} for Phoenix, San Francisco, and Los Angeles were combined proportional to the miles driven in the Waymo RO service, as shown in Table~\ref{tab:human_benchmarks_mileage_blended}. This mileage blended average was compared with the ADS crashed vehicle rate aggregated in the three locations. For the purposes of confidence intervals, the mileage blended benchmark used 19,002 million VMT, which is the average benchmark VMT proportional to the miles driven in the Waymo RO service.

\begin{table*}[width=0.85\textwidth, cols=3]
  \caption{Waymo Mileage Blended Crashed Vehicle Rates and VMT for Human Benchmarks from \citet{scanlon2023benchmark}}
    \begin{tabular*}{\tblwidth}{|l|l|c|}
    \cline{1-3}
    \textbf{Outcome Group} & \textbf{Human Benchmark Source} & \textbf{Human Benchmark IPMM} \\
    \cline{1-3}
    \textit{Any Property Damage or Injury} & Blincoe-Adjusted & 9.67 \\
    \cline{1-3}
    \textit{Police-Reported} & Observed \textit{Police-Reported} & 4.68 \\
    \cline{1-3}
    \multirow{2}{*}{\textit{Any-Injury-Reported}} & Observed & 1.92 \\
    \cline{2-3}
    & Blincoe-Adjusted & 2.80\\
    \cline{1-3}
    \end{tabular*}
  \label{tab:human_benchmarks_mileage_blended}
\end{table*}

The second comparison made in this study was between ADS NHTSA SGO crashes with a police report filed and the observed human police-reported rate. Police-reported and alleged-injury crashes were identified from the SGO-reported crashes using fields in the NHTSA SGO data. The human police-reported rates in \citet{scanlon2023benchmark} did not have an underreporting correction applied. Third, ADS NHTSA SGO crashes where any injuries were reported were compared to the human injury benchmark from \citet{scanlon2017injury}, both adjusted for underreporting using \citet{blincoe2022economic} and unadjusted. Because \textit{police-reported} and \textit{any-injury-reported} crashes are relatively rare, most human NDS-based benchmarks have few of these crashes and do not attempt to estimate rates. Therefore, the only comparable benchmark data sources available were the police reported and any injury reported rate estimates presented in \citet{scanlon2023benchmark}.

This study did not compare the Waymo ADS performance to all human benchmarks developed in \citet{scanlon2023benchmark}. \citet{scanlon2023benchmark} also presented an \textit{any property damage or injury} crash rate using a property damage only underreporting estimated from the SHRP-2 NDS developed in \citet{blanco2016automated}. This current study did not compare the ADS to this Blanco-adjusted benchmark because \citet{blanco2016automated} only attempted to estimate unreported crashes (those that met the reporting threshold but were not reported) where the \citet{blincoe2022economic} underreporting amount accounted for both unreported crashes and crashes that did not meet the reporting threshold \citep{scanlon2023benchmark}. \citet{scanlon2023benchmark} also presented a tow-away benchmark (i.e., any vehicle was towed from the scene due to damage). This study did not compare the ADS to this tow-away benchmark because ADS vehicles are likely towed from the scene more often than human vehicles for reasons that are unrelated to the amount of damage caused by the crash (e.g., due to damage to sensor covers). Finally, the ADS was not compared to the airbag deployed, suspected serious injury+, or fatal injury benchmarks because there are not yet sufficient miles in the ADS data to draw conclusions about these more rare outcomes (see statistical power analysis of \citet{kalra2016driving} and \citet{scanlon2023benchmark}, discussed further in Appendix~\ref{apdx:serious_and_fatal} of the current paper)

Details of the ADS data selection criteria, the collision model, and a list of all crash incidents included for analysis in this paper is included in Appendix~\ref{apdx:selection_of_sgo}.

\subsection*{Confidence Intervals}

Two types of confidence intervals are presented in this paper. First, confidence intervals are estimated for incidents per million miles (IPMM) using a Poisson exact model. The confidence interval, IPMM, can be computed as 
\begin{equation}
IPM \in [ qgamma(\alpha/2, n, 1) / m, qgamma(1 - \alpha/2, n + 1, 1) / m ] 
\label{eqn:ipmm_ci}
\end{equation}
where $qgamma$ is the quantile function of the gamma distribution (i.e., inverse cumulative distribution function), $\alpha$ is the significance level (e.g., 0.05), $n$ is the event count, and $m$ is the number of miles driven. Second, confidence intervals for the ratio between the ADS crash rate of $Y$ events over $t$ miles and the benchmark crash rate of $X$ events over $s$ miles were calculated using the method described by \citet{nelson1970confidence} as
\begin{equation}
 \rho \in [s / t * qbetaprime(\alpha / 2, Y, X+1) , s / t * qbetaprime(1 -\alpha/2, Y+1, X) ] 
 \label{eqn:ratio_ci}
\end{equation}
where $qbetaprime$ is the quantile function of the beta prime distribution. This formulation of the confidence intervals can be computed in the event of zero event counts in the ADS data (i.e., $Y = 0$), but as a result may be more conservative (i.e., wider confidence intervals) in the case of non-zero ADS events counts compared to the Poisson exact model. Appendix~\ref{apdx:nelson_cis} provides example code and test cases.

Another interpretation of the ratio between ADS rate ($\mu = Y/t$) and benchmark rate ($\lambda = X / s$) is a relative difference in the rate, $\delta$, 
\begin{equation}
  \delta = (\mu - \lambda) / \lambda = \mu / \lambda - 1 .
  \label{eqn:reduction}
\end{equation}
Because this percent change in the rate is a direct product of the ratio, the same confidence intervals described in Equation~\ref{eqn:ratio_ci} can be used to generate confidence intervals for the relative difference.

The national crash rates from \citet{scanlon2023benchmark} were derived from crash counts from the CRSS database, which features a complex survey design for sampling crashes. Standard errors for counts that use the survey design variables are larger than standard errors calculated using conventional methods using just the weighted counts \citep{zhang2020crss_gen_var}. To investigate the effect of the CRSS variance estimates on the confidence intervals for the rate ratio, the authors constructed confidence intervals using a parametric bootstrap using the standard error for the benchmark crash counts estimated using the survey design variables. The rate ratio confidence intervals computed using the parametric bootstrap method were narrower than those computed using the method described in Equation~\ref{eqn:ratio_ci}. This result suggests that the method from \citet{nelson1970confidence} is more sensitive to the other parameters in the rate ratio calculation (i.e., the count of ADS events) than a parametric bootstrap method. The \citet{nelson1970confidence} method is more conservative (i.e., produces larger confidence intervals) relative to the parametric boostrap method. For this reason, the confidence intervals described in Equation~\ref{eqn:ratio_ci} were used to construct rate ratio confidence intervals for the national benchmark comparisons.

\section*{RESULTS}\label{section:results}

In total, this study considered the first 7.14 million miles of driving by the Waymo RO service, which drove 5.34 million miles in Phoenix, 1.76 million miles in San Francisco, and 0.0467 million (i.e., 46.7 thousand) miles in Los Angeles. Table~\ref{tab:event_counts} shows the number, IPMM, and 95\% confidence intervals for the different event measures examined in this study. Results for Los Angeles, which are based on a single collision in the \textit{any property damage or injury} category, are listed in Appendix~\ref{apdx:la_results}.

\begin{table*}[width=0.85\textwidth, cols=7]
  \caption{Number and Incidents per Million Miles (IPMM) of Crashed Vehicles for Waymo NHTSA SGO-Reported Crashes with 95\% confidence intervals in Phoenix and San Francisco.}
  \label{tab:event_counts}
  \begin{tabular}{|l|c|c|c|c|c|c|}
    \hline
    \multicolumn{1}{|c|}{} & \multicolumn{3}{c|}{\textbf{Phoenix}} & \multicolumn{3}{c|}{\textbf{San Francisco}} \\
    \cline{2-7}
    \textbf{Measure} & \textbf{n} & \textbf{IPM} & \textbf{CI} & \textbf{n} & \textbf{IPM} & \textbf{CI} \\
    \hline
    RO Miles (millions)& 5.34 & - & - & 1.76 & - & - \\
    \hline
    SGO-Reported & 38 & 7.1 & (5.0, 9.8) & 34 & 19.4 & (13.4, 27.1) \\
    \hline
    SGO-Reported in Transport & 33 & 6.2 & (4.3, 8.7) & 29 & 16.5 & (11.1, 23.7) \\
    \hline
    SGO-Reported Exclude Low Delta-V & 17 & 3.2 & (1.9, 5.1) & 14 & 8.0& (4.4, 13.4) \\
    \hline
    SGO Police-Reported & 12 & 2.2 & (1.2,3.9) & 3 & 1.7 & (0.4, 5.0) \\
    \hline
    SGO \textit{any-injury-reported} & 3 & 0.6 & (0.1, 3.2) & 1 & 0.6 & (<0.1, 3.2) \\
    \hline
  \end{tabular}
\end{table*}

Table~\ref{tab:result_any_prop} and \ref{tab:result_any_prop_exclude_low_severity} report the rate ratio of ADS to human crashed vehicle rates and 95\% confidence intervals for the \textit{any property damage or injury} outcome group. Results with bold print and an asterisk indicate statistical significance (i.e., the 95\% confidence intervals do not contain a 1.0 rate ratio). The human benchmarks are compared to two subsets of the ADS crash data: all NHTSA SGO-reported crashes (Table~\ref{tab:result_any_prop}) and SGO-reported crashes excluding low delta-V (< 1 mph) (Table~\ref{tab:result_any_prop_exclude_low_severity}). The former is the most broad definition of the any property damage without any lower threshold for property damage. The latter is shown to demonstrate the sensitivity of the results to the minimum inclusion criteria on the ADS side, as the human benchmarks all had different inclusion criteria for \textit{any property damage or injury}. Of the 63 SGO-reported in transport crashes, 32 crashes (or 51\%) had a delta-V greater than 1 mph.

\begin{table*}[width=\linewidth, cols=7]
  \caption{Comparison of Waymo ADS and Human Benchmark Crashed Vehicle Rate for Crashes with Any Property Damage or Injury - All NHTSA SGO-Reported Crashes.}
  \label{tab:result_any_prop}
  \begin{tabular}{|p{2.0in}|p{1.4in}|p{0.4in}|p{0.35in}|p{0.45in}|p{0.25in}|p{0.25in}|}
  \cline{1-7}
  \textbf{Human Benchmark} & \textbf{Location} & \textbf{Human IPMM} & \textbf{ADS IPMM} & \textbf{ADS to Human Rate Ratio} &  \multicolumn{2}{c|}{\parbox[t][][t]{0.8in}{\textbf{95\% Percent Confidence Intervals}}} \\
  \cline{1-7}
  \multirow{4}{*}{\parbox{2.0in}{\raggedright \citet{scanlon2023benchmark} Blincoe-adjusted}} & Phoenix & 9.43 & 6.2 & \textbf{0.65*} & \textbf{0.43} & \textbf{0.96} \\
  \cline{2-7} 
  &  San Francisco & 10.5 & 16.5 & 1.57 & 0.99 & 2.37 \\
  \cline{2-7}
  & \raggedright Total - Mile Blend & 9.67 & 8.8 & 0.91 & 0.67 & 1.20 \\
  \cline{2-7}
  & \raggedright Total - National Average & 8.94 & 8.8 & 0.99 & 0.72 & 1.30 \\
  \cline{1-7}
  \hline
  \raggedright \citet{flannagan2023establishing} Ride-hailing NDS & San Francisco & 64.9 & 19.4 & \textbf{0.30*} & \textbf{0.19} & \textbf{0.44} \\
  \hline
  \raggedright \citet{blanco2016automated} SHRP-2 NDS & All Locations & 20.2 & 10.2 & \textbf{0.51*} & \textbf{0.38} & \textbf{0.67} \\
  \cline{1-7}
  \end{tabular}
  \raggedright
  \footnotesize{* and \textbf{bold text} indicates statistically significant at 95\% confidence.}
\end{table*}

\begin{table*}[width=\linewidth, cols=7]
  \caption{Comparison of Waymo ADS and Human Benchmark Crashed Vehicle Rate for Crashes with Any Property Damage or Injury - SGO-Reported Crashes Excluding Low Delta-V.}
  \label{tab:result_any_prop_exclude_low_severity}
  \begin{tabular}{|p{2.0in}|p{1.4in}|p{0.4in}|p{0.35in}|p{0.45in}|p{0.25in}|p{0.25in}|}
  \cline{1-7}
  \textbf{Human Benchmark} & \textbf{Location} & \textbf{Human IPMM} & \textbf{ADS IPMM} & \textbf{ADS to Human Rate Ratio} &  \multicolumn{2}{c|}{\parbox[t][][t]{0.8in}{\textbf{95\% Confidence Intervals}}} \\
  \cline{1-7}
  \multirow{4}{*}{\parbox{2.0in}{\raggedright \citet{scanlon2023benchmark} Blincoe-adjusted}} & Phoenix & 9.43 & 3.2 & \textbf{0.34*} & \textbf{0.18} & \textbf{0.57} \\
  \cline{2-7} 
  &  San Francisco & 10.5 & 8.0 & 0.76 & 0.38 & 1.36 \\
  \cline{2-7}
  & \raggedright Total - Mile Blend & 9.67 & 4.5 & \textbf{0.46*} & \textbf{0.30} & \textbf{0.68} \\
  \cline{2-7}
  & \raggedright Total - National Average & 8.94 & 4.5 & \textbf{0.50*} & \textbf{0.32} & \textbf{0.74} \\
  \cline{1-7}
  \hline
  \raggedright \citet{flannagan2023establishing} Ride-hailing NDS & San Francisco & 64.9 & 8.0 & \textbf{0.12*} & \textbf{0.06} & \textbf{0.22} \\
  \cline{1-7}
  \raggedright \citet{blanco2016automated} SHRP-2 NDS & All Locations & 20.2 & 4.5 & \textbf{0.22*} & \textbf{0.14} & \textbf{0.33} \\
  \cline{1-7}
  \end{tabular}
  \raggedright
  \footnotesize{* and \textbf{bold text} indicates statistically significant at 95\% confidence.}
\end{table*}

\begin{figure}
    \centering
    \includegraphics[width=3in, keepaspectratio]{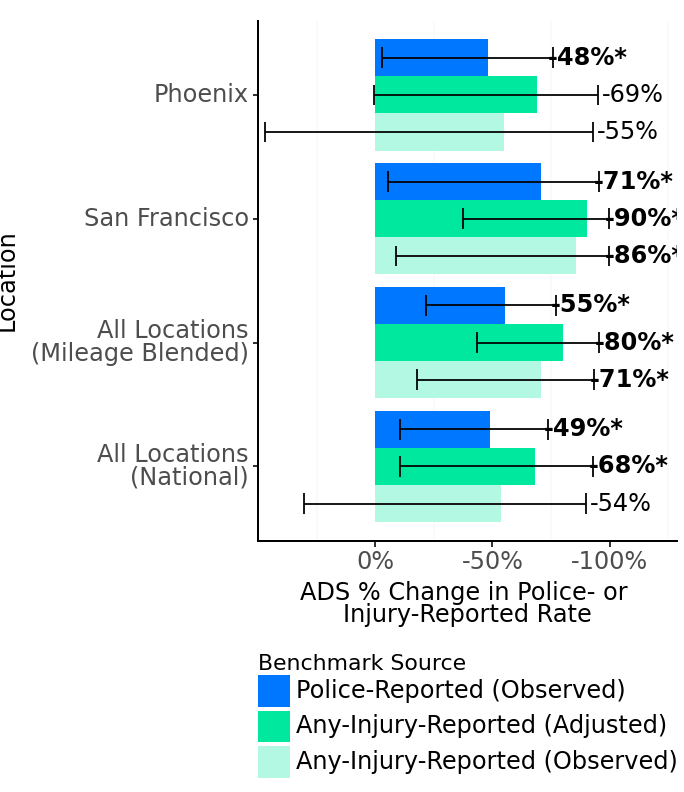}
    \caption{
      Percent Reduction for the Waymo ADS in \textit{Police-Reported} and
      \textit{Any-Injury-Reported} Crashes Compared to a Human Benchmark from
      \citet{scanlon2023benchmark} in Phoenix, San Francisco, and All Locations.
      Note that Los Angeles data is not plotted because of large error bars. See the results in the Appendix}
    \label{fig:percent_reduction_pr_inj}
\end{figure}

Table~\ref{tab:result_inj_pr} shows the ADS to Human rate ratio and 95\% confidence intervals for \textit{police-reported} and \textit{any-injury-reported} crashed vehicle rates. The ADS had a lower \textit{police-reported} and Blincoe-adjusted \textit{any-injury-reported} crashed vehicle rate. The comparison was statistically significant in all locations for the \textit{police-reported} outcome and in San Francisco and when considered in aggregate for the \textit{any-injury-reported} outcome. The observed \textit{any-injury-reported} comparison was not statistically significant in Phoenix and when compared to a national average, but significant in San Francisco and in the mileage blended comparison. The comparison of the ADS to the observed \textit{any-injury-reported} benchmark serves as an analysis of the sensitivity of the results to the Blincoe injury underreporting estimate. The results show that even when not accounting for known underreporting in human data, the point estimates still show the ADS had a lower \textit{any-injury-reported} crash rate, although not all comparisons were statistically significant due to the low event counts and mileage. Inverting the rate ratios in Table~\ref{tab:result_inj_pr}, humans have between 1.9 and 3.4 times the rate of \textit{police-reported} crashes and between 3.1 and 10 times the rate of Blincoe-adjusted \textit{any-injury-reported} crashes as the Waymo ADS. 
When considering all locations together, the \textit{any-injury-reported} crashed vehicle rate was 0.6 incidents per million miles (IPMM) for the ADS vs 2.8 IPMM for the human benchmark, an 80\% reduction or a human crash rate that is 5 times higher than the ADS rate. \textit{Police-reported} crashed vehicle rates for all locations together were 2.1 IPMM for the ADS vs. 4.68 IPMM for the human benchmark, a 55\% reduction or a human crash rate that was 2.2 times higher than the ADS rate.
Figure~\ref{fig:percent_reduction_pr_inj} shows the percent reduction of the ADS for \textit{police-reported} and \textit{any-injury-reported} crashed vehicle rates and 95\% confidence intervals. This percent reduction is another common form of the rate ratios shown in Table~\ref{tab:result_inj_pr}.

\begin{table*}[width=\linewidth, cols=8]
  \caption{Comparison of ADS and Human Benchmark Crashed Vehicle Rate for Police Reported and Any Injury Crashes.}
  \label{tab:result_inj_pr}
  \begin{tabular}{|p{1.5cm}|p{1.4in}|p{1.2in}|p{0.4in}|p{0.35in}|p{0.45in}|p{0.25in}|p{0.25in}|}
  \hline
  \textbf{ADS Events} & \textbf{Human Benchmark} & \textbf{Location} & \textbf{Human IPMM} & \textbf{ADS IPMM} & \textbf{ADS to Human Rate Ratio} & \multicolumn{2}{c|}{\parbox[t][][t]{0.8in}{\textbf{95\% Confidence Intervals}}} \\
  \hline
  \multirow{4}{1.5cm}{\textit{Police-Reported}} & \multirow{4}{*}{\parbox[t][][t]{1.4in}{\citet{scanlon2023benchmark} Observed \textit{Police-Reported}}} & Phoenix & 4.31 & 2.2 & \textbf{0.52}* & \textbf{0.24} & \textbf{0.97} \\
  \cline{3-8}
  & & San Francisco & 5.86 & 1.7 & \textbf{0.29*} & \textbf{0.05} & \textbf{0.95} \\
  \cline{3-8}
  & & Total Mileage Blended & 4.68 & 2.1 & \textbf{0.45*} & \textbf{0.23} & \textbf{0.78} \\
  \cline{3-8}
  & & Total National Average & 4.1 & 2.1 & \textbf{0.51*} & \textbf{0.26} & \textbf{0.90} \\
  \hline
  \multirow{8}{1.5cm}{\textit{Any-Injury-Reported}}  & \multirow{4}{1.4in}{\citet{scanlon2023benchmark} Observed \textit{Any-Injury-Reported}} & Phoenix & 1.24 & 0.6 & 0.45 & 0.07 & 1.47 \\
  \cline{3-8}
  & & San Francisco & 3.98 & 0.6 & \textbf{0.14*} & \textbf{0.002} & \textbf{0.91} \\
  \cline{3-8}
  & & Total Mileage Blended & 1.91 & 0.6 & \textbf{0.29*} & \textbf{0.06} & \textbf{0.82} \\
  \cline{3-8}
  & & Total National Average & 1.21 & 0.6 & 0.46 & 0.10 & 1.30 \\
  \cline{2-8}
  & \multirow{4}{1.4in}{\citet{scanlon2023benchmark} Blincoe-adjusted \textit{Any-Injury-Reported}} & Phoenix & 1.81 & 0.6 & 0.31 & 0.05 & 1.01 \\
  \cline{3-8}
  & & San Francisco & 5.82 & 0.6 & \textbf{0.10*} & \textbf{0.001} & \textbf{0.62} \\
  \cline{3-8}
  & & Total Mileage Blended & 2.80 & 0.6 & \textbf{0.20*} & \textbf{0.04} & \textbf{0.56} \\
  \cline{3-8}
  & & Total National Average & 1.76 & 0.6 & \textbf{0.32*} & \textbf{0.07} & \textbf{0.90} \\
  \hline
  \end{tabular}
  
  \raggedright
  \footnotesize{* and \textbf{bold text} indicates statistically significant at 95\% confidence.}
\end{table*}

\section*{DISCUSSION}\label{section:discussion}

\subsection*{Comparison to Previous Studies}

This study is one of the first to compare RO ADS crash rates to comparable benchmarks in the geographic locations in which ADS currently operate. In addition to comparing \textit{any property damage or injury} outcomes from SGO-reported crashes, this paper also compares \textit{police-reported} and \textit{any-injury-reported} subsets of the SGO data to human benchmarks. 

Many previous studies compared early ADS testing data from California, where a human behind the wheel supervised the ADS, sometimes referred to as Testing Operations (TO) \citep{banerjee2018hands, blanco2016automated, dixit2016autonomous, favaro2017cadmv, goodall2021comparison, schoettle2015preliminary, teoh2017rage}. All of these early studies use data prior to 2020 when RO operations started and thus the results from the current study may be more relevant in understanding the current state of the art of RO deployments.

Several studies have examined either more recent TO data or RO data. The human benchmark derived by \citet{flannagan2023establishing} from ride-hailing NDS was compared to the first 1 million miles of RO driving by the company Cruise, which also operated an RO ride-hailing in San Francisco \citep{zhang_cruise_1M_RO}. \citet{zhang_cruise_1M_RO} found a 65\% reduction in the \textit{any property damage or injury} outcome category, compared to a 70\% reduction found in the current study (0.30 rate ratio in San Francisco compared to the \citet{flannagan2023establishing} benchmark). \citet{chen2024} compared both testing operations and RO crash rates in San Francisco to a human ride-hailing driver crash rate collected as part of data reporting required in California by the California Public Utility Commission (CPUC). The reporting requirement for this CPUC human ride-hailing crash data is not clearly specified, but presumably the requirement is close to the \textit{any property damage or injury} threshold. \citet{chen2024} found the human ride-hailing crash rate was 15.5 IPMM in 2020, compared to an 14.1 IPMM for just under 1 million miles of Waymo RO vehicles from September 2022 to August 2023, which represents the ADS reducing the crash rate by 9\% compared to the benchmark. No statistical tests were performed by \citet{chen2024} due to the low mileage in the ADS data. The human ride-hailing benchmark in San Francisco reported by \citet{chen2024} is higher than the \citet{scanlon2023benchmark} Blincoe-adjusted \textit{any property damage or injury} benchmark and lower than the \citet{flannagan2023establishing} ride-hailing NDS benchmark. Due to an unclear lower reporting threshold in the CPUC data used to derive the human ride-hailing benchmark in \citet{chen2024}, it is difficult to interpret the discrepancies between the \citet{chen2024} and \citet{flannagan2023establishing} benchmarks because the unclear reporting mechanisms in the CPUC data and their implications on underreporting in the human benchmark.

\citet{cummings2024assessing} compared SGO-reported collisions in California from 2021 to 2022 to a police-reported crash rate. The data used in \citet{cummings2024assessing} mixed RO and TO data from California. \citet{cummings2024assessing} compared a Waymo ADS \textit{any property damage or injury} crash (all SGO-reported crashes) to a \textit{police-reported} crash rate derived from national crash and VMT data. As discussed previously, the SGO reporting threshold includes any property damage with no lower reporting threshold, whereas police-reported data requires higher amount of property damage or an injury. Therefore, comparing the ADS SGO crash rate to a police-reported crash rate introduces selection bias that deflates the human benchmark relative to the ADS data. Furthermore, the effects of underreporting in the human data were not accounted for in \citet{cummings2024assessing}, introducing reporting bias that also deflates the human crash rate relative to the ADS crash rate. Due to these biases, \citet{cummings2024assessing} incorrectly concluded that the Waymo ADS had a statistically significant 4-times-higher crash rate than the human benchmark. Indeed, when comparing the Waymo RO \textit{any property damage or injury} rate in San Francisco from \citet{cummings2024assessing} (16.5 IPMM) to the national \citet{scanlon2023benchmark} observed \textit{police-reported} benchmark (4.10 IPMM), the ADS mistakenly appears to have a 4.0-times-higher crash rate. The \textit{any property damage or injury} and \textit{police-reported} comparisons done in the current paper control for these selection and reporting biases by using benchmarks that attempt to match the reporting threshold and account for underreporting, and additionally use benchmarks restricted to passenger vehicles and from the areas the ADS operates in (the \textit{police-reported} crash rate in San Francisco of 5.86 IPMM is 40\% larger than the national average of 4.10 IPMM).

The benchmarks and ADS data presented in this paper are overall crash involvement rates which do not consider crash contribution of those involved. \citet{di2023comparative} compared third party liability property damage and bodily injury insurance claims rates between the Waymo ADS operations and a human benchmark. A liability claim is a request for compensation when someone is
responsible for damage to property or injury to another person, typically following a collision. In \citet{di2023comparative}, only 3rd party liability claims that were resolved or were likely to be resolved with a payment were included as these types of claims are a proxy for responsibility. \citet{di2023comparative} found that over 3.8 million miles of RO driving, the Waymo RO service had a statistically significant, lower property damage and bodily injury claims rate compared to a human benchmark based on aggregate human insurance claims data restricted to zip codes where the Waymo service operates. Examining third party liability claims allows an analysis of the degree to which the ADS contributes to crash causation, that is, the ADS safety impact on other road users. The results of the current study, in contrast, examine the overall crashed vehicle rate regardless of the contribution to the crash’s causation. Many of the crashes in the RO dataset examined in this paper featured the Waymo ADS parked or stopped appropriately at a traffic control device where the ADS’s behavior had little or no contribution to the crash’s cause. Examining ADS performance from both a crash causation and overall crash rate is important because it provides assessment of (a) performance in crashes where there was some contribution (3rd party liability claims), (b) performance in crashes where there was no contribution, and (c) the combination (overall crash rates). \citet{di2023comparative} reported that Waymo RO operations reduced property damage liability claims by 76\% and bodily injury liability claims by 100\%. This magnitude of claims reduction is similar to the magnitude in \textit{police-reported} (49\% to 71\% reduction) and \textit{any-injury-reported} (69\% and 90\% reduction) reductions reported in the current study, which did not consider contribution to causing the crash.

\subsection*{Interpretation of Results}

\subsubsection*{Any Property Damage or Injury}

The available benchmark crashed vehicle rates for \textit{any property damage or injury} crashes vary widely depending on the method of estimation (i.e., underreporting adjustment of crash data vs. observations in NDS) and the driving environment (e.g., SF ride-hailing NDS vs. national average NDS). Both NDS benchmarks used a four-level severity scale from the SHRP-2 NDS \citep{hankey2016description} and include level 1 through level 3 severity crashes in their benchmark \citet{flannagan2023establishing, blanco2016automated}. In particular, the most frequent level 3 events include events like road departures and curb strikes with an ``increased risk element'' without vehicle damage, as well as crashes resulting in minor damage. This level 3 definition does not match the SGO-reporting requirement of property damage nor does it likely match the various crash definitions used to construct the underreporting adjustment in \citet{blincoe2022economic}. This may partially explain why the NDS crash rates are higher than the Blincoe-adjusted rates. Even still, given the available data it is unclear what magnitude of effect on the \textit{any property damage or injury} benchmark is due to unclear or different reporting thresholds, underreporting differences by location, or whether the different characteristics of the driving populations (all drivers vs ride-hail drivers using rental vehicles). For example, \citet{scanlon2023benchmark} found that state data from California had a much lower proportion of reported property damage only crashes relative to injury crashes compared with Arizona and national data, which strongly suggests the national \citet{blincoe2022economic} may not generalize to individual states. \citet{blanco2016automated} attempted to estimate the rate of collisions that met a reporting threshold but were not reported to police, which resulted in a 84\% property damage only underreporting amount compared to a 60\% underreporting found by \citet{blincoe2022economic}. Although the \citet{blanco2016automated} study had fewer participants and other limitations,  the discrepancy suggests there is potential variability in the property damage underreporting amount.

A conservative interpretation of the results of the \textit{any property damage or injury} outcome group is that conclusions cannot be drawn for this outcome group until  benchmarks are further developed. The comparisons in the current study do show, however, the possibility of the ADS reducing crash rates. The ADS reduced crash rates in the comparison of all SGO crashes to a Blincoe-adjusted police-reported rate in Phoenix at a statistically significant level, but a higher (but not statistically significant) crash rate for the Blincoe-adjusted rate in San Francisco. The lower property damage only reporting in California discussed above suggests that the true crash rate in San Francisco may be higher than was estimated from the police-reported crash data, but to what degree could not be determined. The comparisons of all SGO collisions to the NDS benchmarks from \citet{flannagan2023establishing} and \citet{blanco2016automated} showed a large, statistically significant reduction in crash rate for the ADS. As discussed above, however, there is likely events that do not result in property damage in the NDS data samples (i.e., the ``level 3'' events from the SHRP-2 event definitions), and thus the NDS crash rates are likely inflated to some degree relative to the ADS crash rate. The comparison of SGO collisions excluding crashes with delta-V less than 1 mph showed that many of the ADS collisions are of low severity. As a result this comparison showed the ADS had a reduction in crash rate at a statistically significant level in all locations except for San Francisco (which again could be partially explained by the property damage reporting discrepancies in California). 

To address these discrepancies and difficulties for the \textit{any property damage or injury} benchmarks, there is a need for future research that can more accurately define an objective lower reporting threshold in human crash data than exists today. Such benchmarks would also require standardized methods for reporting the severity of ADS crashes, like the delta-V threshold used in this study.
The availability of sensor data enables accurate reconstructions of ADS crashes to estimate their severity (e.g., delta-V or impact speed). Estimating crash severity is often not possible from most human crash databases without detailed reconstructions. There are challenges in generating benchmarks for a specific driving environment using sampled or non-representative database that do have such crash reconstructions. Another potential source of severity benchmark is to estimate severity from NDS data. Advances in computer vision may enable crash reconstructions from limited vehicle sensor and video data (see for example \citet{campolettano2023representative}).

\subsubsection*{Police-Reported and Any-Injury-Reported}

Observed \textit{police-reported} crashes, in contrast to \textit{any property damage or injury} crashes, may be a comparison with less inherent systematic variability. Although reporting thresholds and reporting practices may differ by jurisdiction \citep{scanlon2023benchmark}, if observed police reports in the ADS and human benchmark are from the same jurisdictions and time periods, then the reporting practices should be comparable in both populations. 

This comparison of \textit{police-reported} crashed vehicle rates is likely a conservative comparison for several reasons. Private citizens often do not report police-reportable events \citep{mdavis2015survey}. In California, only injury crashes are required to be reported to the state databases, which means some jurisdictions in California may have different PDO police-reported crashes reporting practices. Conversely for ADS crashes, this study conservatively used the ``Law Enforcement Investigating?'' field of the NHTSA SGO reports because the authors have knowledge that Waymo’s operational policies in which it is likely police took information that could be used to file a report or indicated they would file a report. This field, as Waymo reports to the SGO, is only reported ``yes'' when law enforcement indicate they will file a report, not only when law enforcement responds to the scene of a crash. Although the \textit{police-reported} outcome is often associated with a certain level of property damage, factors other than property damage may influence whether an ADS fleet operator report crashes to the police. Future research should investigate police reporting differences between businesses (like an ADS fleet operator) and the general population.

The \textit{Any-injury-reported} outcome group may also serve as a useful comparison between ADS and a benchmark. It has been shown that there is less underreporting in injury crashes compared to crashes with no injuries \citet{blincoe2022economic}. Because the magnitudes of underreporting are smaller, the effect of the underreporting estimate on the comparisons may also be lower, as shown by the results of this study. The injury rates, however, can also suffer from similar surveillance biases as the property damage and police report data related to underreporting. The minimum threshold of an injury is often unclear and is dependent on self-reported injury data. Most injuries in the ADS dataset were minor and did not require medical treatment. As more RO miles are driven, and rarer injury crashes that require medical treatment occur, there is an opportunity to use coded injuries scales, like the Abbreviated Injury Severity (AIS) \citep{aaam2016}, to more precisely define injury thresholds.

\subsection*{Other Limitations}\label{section:limitations}

In addition to the limitations discussed above, this study has several additional limitations. The results in this study examine crashed vehicle rates aggregated by operating location and do not attempt to evaluate crashed vehicle rates within subgroups, for example, by crash partner or crash type. Past studies that reported ADS crash data have found that ADS vehicles may have different proportions of different crash types than the general population \citep{victor2023_1mRO, teoh2017rage}. A lower aggregate crash rate does not preclude an ADS from having an elevated crash rate in certain types of crashes. As more RO data is collected and statistical power allows, future research should investigate crash rates of more granular crash types.

Although the benchmarks used in this study could adjust for some factors that are related to crashed vehicle rate (e.g., geographical area, types of roads, vehicle type), other pertinent factors were not controlled for. These include time of day adjustments and adjustments for the driving density within the geographic areas. As the Waymo ADS is being used as a ride-hailing service, the routes the ADS drives are likely differently dispersed than the general driving population and may be biased toward more densely populated areas. There are challenges when matching benchmark crash and VMT data sources that can be adjusted by these parameters, particularly in VMT data. An aggregate-level analysis comparing the ADS to the current crash status quo should include the ADS route choices, even if different from comparable human drivers, if the research question (like the one posed in this paper) is to determine the safety impact of replacing human driving with ADS vehicle driving.

An ADS vehicle operating in a ride hailing context is not always occupied by a human (e.g., before or in between picking up passengers). Therefore, when examining injury outcomes, if the ADS vehicle is unoccupied, there would be a reduced risk of injury for a similar human-only crash. As in the route choice discussion above, the objective of this study was to compare the ADS crash involvement to current human drivers, so it is appropriate to include miles where the ADS vehicle is not occupied. More broadly, risk of injury in a motor vehicle crash can be affected by many factors, such as seat belt use and vehicle model year (as a proxy for available safety systems). A risk-based approach that uses an injury risk function to compute probability of an injury outcome regardless of the whether an occupant is in the vehicle given some set of inputs can be advantageous in normalizing risk in both the benchmark and ADS crashes for comparison, as has been traditionally done in prospective safety benefit analyses (for example, \citet{kusano2012forwardcrash}; \citet{scanlon2017injury}).

The Waymo RO service studied in this paper was operating over multiple years on different vehicle platforms and with different software versions. Just as the human driving population has changing characteristics over time, so do certain aspects of the Waymo RO ride-hailing service. The operating environment and territory have expanded during this time period. Starting in the area around Chandler, Arizona in 2020, today the Waymo RO ride-hailing service operates in an area that includes multiple cities in metropolitan Phoenix and over almost the entire area of San Francisco. A limitation of this study is that the entire history of the Waymo RO crash record was compared to human crash data over different time periods. As ADS fleets continue to increase the number of VMT, additional analyses that are restricted to certain periods of time or platforms could be possible. Many past retrospective studies of safety technologies, however, group different implementations of the same technology over long time periods to investigate the overall effect of a technology. For example, vehicles from different manufacturers were compared to investigate the effectiveness of first- and second-generation airbags \citet{olson2006association} and electronic stability control \citet{riexinger2019has}. Thus, aggregate performance of RO ADS are still informative at this point in time.

\section*{CONCLUSIONS}\label{section:conclusions}

This study compared the crashed vehicle rate of the Waymo RO service to human benchmarks for \textit{any property damage or injury}, \textit{police-reported}, and \textit{any-injury-reported} outcomes. The results show that the Waymo RO service had a lower \textit{police-reported} and \textit{any-injury-reported} crashed vehicle rate compared to the benchmarks. 
When considering all locations together, the \textit{any-injury-reported} crashed vehicle rate was 0.6 incidents per million miles (IPMM) for the ADS vs 2.8 IPMM for the human benchmark, an 80\% reduction or a human crash rate that is 5 times higher than the ADS rate. \textit{Police-reported} crashed vehicle rates for all locations together were 2.1 IPMM for the ADS vs. 4.68 IPMM for the human benchmark, a 55\% reduction or a human crash rate that was 2.2 times higher than the ADS rate.
The ADS \textit{any-injury-reported} and \textit{police-reported} crashed vehicle rates lower than the benchmark rates at a statistically significant level over a total of 7.14 million RO miles when compared individually in Phoenix and San Francisco and when compared overall (except for the \textit{any-injury-reported} comparison in Phoenix).
Although these results are promising, they are based on a small number of ADS events (15 police-reported and 4 any injury-reported). Thus, these comparisons may be subject to statistical noise as more miles and events are observed in the future.
The comparison for Los Angeles was not statistically significant for zero observed ADS events but few miles. The results for the \textit{police-reported} and \textit{any-injury-reported} outcome groups, however, show a positive safety impact of the ADS in these types of crashes.

The results are more difficult to interpret for the \textit{any property damage or injury} outcome comparison. The direction of the result and the statistical significance of the comparison depended on which benchmark was used (under-reporting adjusted police-report or NDS data) and whether low delta-V (< 1 mph) crashes were excluded from the ADS comparison. The results of the \textit{any property damage or injury} comparison showed a statistically significant decrease in crashes in Phoenix when the ADS was compared to a police-reported benchmark adjusted for underreporting and in San Francisco and overall when compared to NDS benchmarks. There was an increase in the ADS crash rate when compared to an underreporting adjusted benchmark in San Francisco that was not statistically significant. Half of all ADS crashes had a delta-V less than 1 mph. When excluding ADS crashes with a delta-V less than 1 mph, the decrease in \textit{any property damage or injury} crash rate for the ADS was statistically significant in all comparisons except in the comparison to the underreporting adjusted benchmark from San Francisco. These inconclusive results are largely due to an unclear lower reporting threshold in the human benchmarks. More research is needed to develop benchmarks that have an objective lower reporting threshold to draw conclusions about the \textit{any property damage or injury} outcome group. 

This is one of the first studies to compare overall crashed vehicle rates of ADS RO data only (as opposed to a mix of ADS testing with a human behind the wheel) to human benchmarks that also corrected for biases reported in the literature. The comparison of crash rates to benchmarks is part of the safety determination lifecycle, which itself is a part of a broader ADS safety case. The results show that the Waymo ADS operating in RO configuration has a lower crashed vehicle rate than human drivers when considering \textit{police-reported} and \textit{any-injury-reported} crashes. This result provides directional confirmation of the predictions and analyses used prior to deploying RO operations.

\section*{DISCLOSURE STATEMENT}
All authors are employed by Waymo LLC.

\bibliographystyle{cse_author_name}
\bibliography{refs}

\begin{thebibliography}{33}
\providecommand{\natexlab}[1]{#1}
\providecommand{\url}[1]{\texttt{#1}}
\providecommand{\urlprefix}{URL }
\expandafter\ifx\csname urlstyle\endcsname\relax
  \providecommand{\doi}[1]{doi:\discretionary{}{}{}#1}\else
  \providecommand{\doi}{doi:\discretionary{}{}{}\begingroup
  \urlstyle{rm}\Url}\fi

\bibitem[{AAAM(2016)}]{aaam2016}
AAAM. 2016.
\newblock The Abbreviated Injury Scale-2015 Revision.

\bibitem[{Banerjee et~al.(2018)Banerjee, Jha, Cyriac, Kalbarczyk, and
  Iyer}]{banerjee2018hands}
Banerjee S, Jha S, Cyriac J, Kalbarczyk ZT, and Iyer RK. 2018.
\newblock Hands off the wheel in autonomous vehicles?: A systems perspective on
  over a million miles of field data.
\newblock In: 2018 48th Annual IEEE/IFIP International Conference on Dependable
  Systems and Networks (DSN), pp. 586--597. IEEE.

\bibitem[{Blanco et~al.(2016)Blanco, Atwood, Russell, Trimble, McClafferty, and
  Perez}]{blanco2016automated}
Blanco M, Atwood J, Russell SM, Trimble TE, McClafferty JA, and Perez MA. 2016.
\newblock Automated vehicle crash rate comparison using naturalistic data.
\newblock Tech. rep., Virginia Tech Transportation Institute.
\newblock \urlprefix\url{https://vtechworks.lib.vt.edu/handle/10919/64420}.

\bibitem[{Blincoe et~al.(2023)Blincoe, Miller, Wang, Swedler, Coughlin,
  Lawrence, Guo, Klauer, and Dingus}]{blincoe2022economic}
Blincoe L, Miller TR, Wang JS, Swedler D, Coughlin T, Lawrence B, Guo F, Klauer
  S, and Dingus T. 2023.
\newblock The economic and societal impact of motor vehicle crashes, 2019
  (revised).
\newblock Tech. Rep. DOT HS 813 403, National Highway Traffic Safety
  Administration.

\bibitem[{Campolettano et~al.(2023)Campolettano, Scanlon, and
  Victor}]{campolettano2023representative}
Campolettano ET, Scanlon JM, and Victor T. 2023.
\newblock Representative pedestrian collision injury risk distributions for a
  dense-urban us odd using naturalistic dash camera data.
\newblock In: Proceedings of the Enhanced Safety of Vehicles Conference,
  23-0075.

\bibitem[{Chen and Shladover(2024)}]{chen2024}
Chen JJ and Shladover SE. 2024.
\newblock Initial indications of safety of driverless automated driving
  systems.
\newblock In: The 103rd Transportation Research Board (TRB) Annual Meeting.
  Washington, D.C.

\bibitem[{Cummings(2024)}]{cummings2024assessing}
Cummings M. 2024.
\newblock Assessing readiness of self-driving vehicles.
\newblock In: The 103rd Transportation Research Board (TRB) Annual Meeting.
  Washington, D.C.
\newblock Preprint.

\bibitem[{Di~Lillo et~al.(2023)Di~Lillo, Gode, Zhou, Atzei, Chen, and
  Victor}]{di2023comparative}
Di~Lillo L, Gode T, Zhou X, Atzei M, Chen R, and Victor T. 2023.
\newblock Comparative safety performance of autonomous-and human drivers: A
  real-world case study of the waymo one service.
\newblock arXiv preprint arXiv:230901206.
\newblock \doi{https://doi.org/10.48550/arXiv.2309.01206}.

\bibitem[{Dixit et~al.(2016)Dixit, Chand, and Nair}]{dixit2016autonomous}
Dixit V, Chand S, and Nair D. 2016.
\newblock Autonomous vehicles: disengagements, accidents and reaction times.
\newblock PLoS one, 11:e0168,054.

\bibitem[{Favar{\`o} et~al.(2023{\natexlab{a}})Favar{\`o}, Fraade-Blanar,
  Schnelle, Victor, Pe{\~n}a, Engstrom, Scanlon, Kusano, and
  Smith}]{favaro2023building}
Favar{\`o} F, Fraade-Blanar L, Schnelle S, Victor T, Pe{\~n}a M, Engstrom J,
  Scanlon J, Kusano K, and Smith D. 2023{\natexlab{a}}.
\newblock Building a credible case for safety: Waymo's approach for the
  determination of absence of unreasonable risk.
\newblock arXiv preprint arXiv:230601917.
\newblock \doi{https://doi.org/10.48550/arXiv.2306.01917}.

\bibitem[{Favar{\`o} et~al.(2017)Favar{\`o}, Nader, Eurich, Tripp, and
  Varadaraju}]{favaro2017cadmv}
Favar{\`o} F, Nader N, Eurich S, Tripp M, and Varadaraju N. 2017.
\newblock Examining accident reports involving autonomous vehicles in
  california.
\newblock PLoS one, 12:e0184,952.

\bibitem[{Favar{\`o} et~al.(2023{\natexlab{b}})Favar{\`o}, Victor, Hohnhold,
  and Schnelle}]{favaro2023interpreting}
Favar{\`o} FM, Victor T, Hohnhold H, and Schnelle S. 2023{\natexlab{b}}.
\newblock Interpreting safety outcomes: Waymo's performance evaluation in the
  context of a broader determination of safety readiness.
\newblock arXiv preprint arXiv:230614923.
\newblock \doi{https://doi.org/10.48550/arXiv.2306.14923}.

\bibitem[{Flannagan et~al.(2023)Flannagan, Leslie, Kiefer, Bogard,
  Chi-Johnston, Freeman, Huang, Walsh, and Anthony}]{flannagan2023establishing}
Flannagan C, Leslie A, Kiefer R, Bogard S, Chi-Johnston G, Freeman L, Huang R,
  Walsh D, and Anthony J. 2023.
\newblock Establishing a crash rate benchmark using large-scale naturalistic
  human ridehail data.
\newblock Tech. Rep. UMTRI-2023-18, University of Michigan Transportation
  Research Institute.
\newblock \urlprefix\url{https://deepblue.lib.umich.edu/handle/2027.42/178179}.

\bibitem[{Goodall(2021)}]{goodall2021comparison}
Goodall NJ. 2021.
\newblock Comparison of automated vehicle struck-from-behind crash rates with
  national rates using naturalistic data.
\newblock Accident Analysis \& Prevention, 154:106,056.

\bibitem[{Hankey et~al.(2016)Hankey, Perez, and
  McClafferty}]{hankey2016description}
Hankey JM, Perez MA, and McClafferty JA. 2016.
\newblock Description of the {SHRP-2} naturalistic database and the crash,
  near-crash, and baseline data sets.
\newblock Tech. rep., Virginia Tech Transportation Institute.

\bibitem[{Kalra and Paddock(2016)}]{kalra2016driving}
Kalra N and Paddock SM. 2016.
\newblock Driving to safety: How many miles of driving would it take to
  demonstrate autonomous vehicle reliability?
\newblock Transportation Research Part A: Policy and Practice, 94:182--193.

\bibitem[{Kusano and Gabler(2012)}]{kusano2012forwardcrash}
Kusano KD and Gabler HC. 2012.
\newblock Safety benefits of forward collision warning, brake assist, and
  autonomous braking systems in rear-end collisions.
\newblock IEEE Transactions on Intelligent Transportation Systems,
  13:1546--1555.
\newblock \doi{https://doi.org/10.1109/TITS.2012.2191542}.

\bibitem[{{\relax M.\ Davis \& Co.}(2015)}]{mdavis2015survey}
{\relax M\ Davis \& Co}. 2015.
\newblock National telephone survey of reported and unreported motor vehicle
  crashes. (findings report).
\newblock Tech. Rep. DOT HS 812 183), National Highway Traffic Safety
  Administration.

\bibitem[{{\relax National Highway Traffic Safety
  Administration}(2023)}]{nhtsa2021sgo}
{\relax National Highway Traffic Safety Administration}. 2023.
\newblock Second amended standing general order 2021-01: Incident reporting for
  automated driving systems (ads) and level 2 advanced driver assistance
  systems (adas).
\newblock
  \urlprefix\url{https://www.nhtsa.gov/sites/nhtsa.gov/files/2023-04/Second-Amended-SGO-2021-01_2023-04-05_2.pdf},
  online; accessed on November 1, 2023.

\bibitem[{Nelson(1970)}]{nelson1970confidence}
Nelson W. 1970.
\newblock Confidence intervals for the ratio of two poisson means and poisson
  predictor intervals.
\newblock IEEE Transactions on Reliability, 19:42--49.
\newblock \doi{https://doi.org/10.1109/TR.1970.5216388}.

\bibitem[{Olson et~al.(2006)Olson, Cummings, and Rivara}]{olson2006association}
Olson CM, Cummings P, and Rivara FP. 2006.
\newblock Association of first-and second-generation air bags with front
  occupant death in car crashes: a matched cohort study.
\newblock American journal of epidemiology, 164:161--169.

\bibitem[{Riexinger et~al.(2019)Riexinger, Sherony, and
  Gabler}]{riexinger2019has}
Riexinger L, Sherony R, and Gabler H. 2019.
\newblock Has electronic stability control reduced rollover crashes?
\newblock Tech. Rep. 2019-01-1022, SAE International.

\bibitem[{Scanlon et~al.(2021)Scanlon, Kusano, Daniel, Alderson, Ogle, and
  Victor}]{scanlon2021fatal}
Scanlon JM, Kusano KD, Daniel T, Alderson C, Ogle A, and Victor T. 2021.
\newblock Waymo simulated driving behavior in reconstructed fatal crashes
  within an autonomous vehicle operating domain.
\newblock Accident Analysis \& Prevention, 163:106,454.
\newblock \doi{https://doi.org/10.1016/j.aap.2021.106454}.

\bibitem[{Scanlon et~al.(2023)Scanlon, Kusano, Fraade-Blanar, McMurry, Chen,
  and Victor}]{scanlon2023benchmark}
Scanlon JM, Kusano KD, Fraade-Blanar LA, McMurry TL, Chen YH, and Victor T.
  2023.
\newblock Benchmarks for retrospective automated driving system crash rate
  analysis using police-reported crash data.
\newblock Traffic Injury Prevention (under review).

\bibitem[{Scanlon et~al.(2017)Scanlon, Sherony, and Gabler}]{scanlon2017injury}
Scanlon JM, Sherony R, and Gabler HC. 2017.
\newblock Injury mitigation estimates for an intersection driver assistance
  system in straight crossing path crashes in the united states.
\newblock Traffic injury prevention, 18:S9--S17.
\newblock \doi{https://doi.org/10.1080/15389588.2017.1300257}.

\bibitem[{Schoettle and Sivak(2015)}]{schoettle2015preliminary}
Schoettle B and Sivak M. 2015.
\newblock A preliminary analysis of real-world crashes involving self-driving
  vehicles.
\newblock Tech. rep., University of Michigan Transportation Research Institute.

\bibitem[{Teoh and Kidd(2017)}]{teoh2017rage}
Teoh ER and Kidd DG. 2017.
\newblock Rage against the machine? google's self-driving cars versus human
  drivers.
\newblock Journal of Safety Research, 63:57--60.

\bibitem[{{\relax United States Department of
  Transportation}(2022)}]{usdot2022nrss}
{\relax United States Department of Transportation}. 2022.
\newblock National roadway safety strategy.
\newblock
  \urlprefix\url{https://www.transportation.gov/sites/dot.gov/files/2022-02/USDOT-National-Roadway-Safety-Strategy.pdf},
  online; accessed on November 1, 2023.

\bibitem[{Victor et~al.(2023)Victor, Kusano, Gode, Chen, and
  Schwall}]{victor2023_1mRO}
Victor T, Kusano K, Gode T, Chen R, and Schwall M. 2023.
\newblock Safety performance of the waymo rider-only automated driving system
  at one million miles.

\bibitem[{Webb et~al.(2020)Webb, Smith, Ludwick, Victor, Hommes, Favar{\`o},
  Ivanov, and Daniel}]{webb2020readiness}
Webb N, Smith D, Ludwick C, Victor T, Hommes Q, Favar{\`o} F, Ivanov G, and
  Daniel T. 2020.
\newblock Waymo's safety methodologies and safety readiness determinations.
\newblock arXiv preprint arXiv:201100054.
\newblock \doi{https://doi.org/10.48550/arXiv.2011.00054}.

\bibitem[{Young(2021)}]{young2021critical}
Young R. 2021.
\newblock Critical Analysis of Prototype Autonomous Vehicle Crash Rates: Six
  Scientific Studies from 2015--2018.
\newblock SAE International.

\bibitem[{Zhang and Diaz(2020)}]{zhang2020crss_gen_var}
Zhang F and Diaz E. 2020.
\newblock Crash report sampling system: Generalized variance functions.
\newblock Tech. Rep. DOT HS 813 041, National Highway Traffic Safety
  Administration.

\bibitem[{Zhang(2023)}]{zhang_cruise_1M_RO}
Zhang L. 2023.
\newblock Human ridehail crash rate benchmark.
\newblock
  \urlprefix\url{https://www.getcruise.com/news/blog/2023/human-ridehail-crash-rate-benchmark/},
  online; accessed November 1, 2023.

\end{thebibliography}

\pagebreak

\appendix

\singlespacing

\section{Appendix}

\subsection{Selection of SGO Events}\label{apdx:selection_of_sgo}

The table below lists all the crashes included in this analysis. The SGO report ID can be used to look up further details of these events that are published by NHTSA. Detailed narratives or other data is not included in this paper, as this information can be obtained from the NHTSA SGO reporting website (\hyperlink{https://www.nhtsa.gov/laws-regulations/standing-general-order-crash-reporting}{https://www.nhtsa.gov/laws-regulations/standing-general-order-crash-reporting}). The ``Waymo 1M Mile RO Event \#'' column lists the integer event identifier that was reported in \citet{victor2023_1mRO} that described all contact events (regardless of whether they were reported as part of the NHTSA SGO). Of the 20 events reported in \citet{victor2023_1mRO}, 11 were reported as part of the NHTSA SGO. Waymo started to operate in RO before the NHTSA SGO was enacted in July 2021 and as a result, 4 contact incidents that were previously reported in \citet{victor2023_1mRO} were not subject to the SGO reporting requirements. The authors retroactively reviewed these 4 contact incidents and determined that 2 of the contact incidents would have qualified for NHTSA SGO reporting. These two (2) incidents were included in the data analyzed for this paper. The location column refers to the market the crash occurred in (PHX for Phoenix, SFO for San Francisco, and LA for Los Angeles). The remaining columns indicate whether the SGO-reported crash belonged to each of the groups analyzed in this paper and described in the methodology section. As noted in the Methods section, two crashes occurred before the NHTSA SGO reporting requirements were enacted but were retroactively reviewed and deemed to meet the SGO reporting requirements (events \#1 and \#4 from \citet{victor2023_1mRO}.

In the SGO data, crashes reported by Waymo were identified using the ``Reporting Entity'' field having the value ``Waymo LLC''. The field ``Driver / Operator Type'' was set to ``None'' for RO operations, and set to ``In-Vehicle (Commercial / Test)'' when an autonomous specialist is monitoring the ADS behind the steering wheel. Note that the 7.1 million miles of ADS vehicle driving that were used to compute the rates presented in the paper only include RO miles. These 7.1 million RO miles should not be used to compute rates for SGO-reported crashes where an autonomous specialist was behind the wheel. The NHTSA SGO requires ADS fleet operators to report crashes where the ADS vehicle contributed or is alleged to have contributed (by steering, braking, acceleration, or other operational performance) to another vehicle’s physical impact with another road user or property involved in a reportable crash even if the ADS vehicle was not involved in the physical impact. Two Waymo RO collisions during the study period involved alleged contribution but no damage to the Waymo vehicle. These two cases are described in more detail in the appendix.

An administrative error in Waymo's SGO submissions prior to February 2022 resulted in some duplicate crash entries in the published NHTSA SGO data. The NHTSA SGO has a facility for updating or making corrections to previously submitted crashes. An updated submission may be made if new information is available that was not available at the time the original report was submitted. Prior to February 2022, Waymo submitted a new SGO report for some crashes that were intended to be updates to previous crashes. This resulted in a new SGO report ID being issued instead of updating the information for the original crash. The event narratives in the published SGO data for these updated reports with new report IDs state they are duplicate submissions that update some information, and reference the original SGO report ID. The following NHTSA report IDs were excluded from this analysis because they are duplicates of updated reports: 30270-2248, 30270-2198, 30270-2168, 30270-2160, 30270-1949, 30270-1815, 30270-1160, 30270-1613, 30270-1748, 30270-1778, 30270-1535.

An ``in-transport'' vehicle is one that is traveling (moving or stopped) in the roadway. ``Not-in-transport'' vehicles are those that are parked in a designated parking spot out of the roadway. Therefore, including ADS vehicle crashes where the ADS vehicle is a not in-transport vehicle would inflate the ADS crash count relative to the benchmark. The ADS vehicle was considered not in-transport if the ADS vehicle was in the ``park'' gear and the ADS vehicle was in a designated parking area. This not in-transport status can happen when the ADS vehicle is parked awaiting its next trip instructions. Because the ADS software is running and in control of the vehicle, contacts while the ADS vehicle is not in-transport are reported as part of the NHTSA SGO reporting requirements. If the parking spot was on-street parking parallel to the traffic flow, the ADS vehicle must have been within 18 inches from the curb of the road as measured by the measured sensor data and curb location in the ADS map to be considered not-in-transport. Instances where the ADS vehicle was in ``park'' gear, but not near the curb were included as in-transport vehicles. The alternative, to include ``not-in-transport'' vehicle ADS crashes, would require also including ``not-in-transport'' vehicles in the benchmark, which are included in separate counts from in-transport vehicles in most police report databases, if at all. Additionally, it is unclear how a rate of crashed vehicles per VMT should be interpreted for not-in-transport vehicles, as parked vehicle crash risk is likely more related to amount of time parked rather than VMT. Underreporting for not-in-transport vehicles may also differ from in-transport vehicles, which was not accounted for the underreporting adjustments used in the benchmarks \citep{scanlon2023benchmark, blincoe2022economic}.

The ``in-transport'' determination was made with ADS sensor and map data that are not part of the crash data published as part of the SGO. The event narratives in the NHTSA SGO data for Waymo can also be used to determine whether the vehicle was parked in a valid parking spot. There were 12 SGO crashes where the Waymo vehicle was in ``park'' gear at the time of the crash. Of those 12, in 7 crashes the Waymo vehile was parked near the curb and in 5 the Waymo vehicle was not near the curb. In 6 out 7 crashes where the Waymo vehicle was near the curb, the event narratives state the vehicle was parked ``near'' or ``adjacent'' to the curb. Of the 8 crashes where the Waymo vehicle was in park but not in a parking spot or near the curb, 2 of the narratives  mentioned the vehicle had pulled ``near'' or ``pulled to'' the curb. Therefore, using the event narratives alone would yield similar results than was found using the additional sensor data (8 in park and near the curb and 7 in park but not near a curb using narratives compared to 7 in park and near the curb and 8 in park but not near a curb when using sensor and map data). There is no reporting requirement in the SGO for parking status nor is there a definition of a ``valid'' parking space. Having agreed upon definitions for ``in-transport'' status would improve comparability between ADS crash data and human crash data.

Two NHTSA SGO-reported crashes did not include damage to the ADS vehicle. First, case 30270-6542 involved two human-driven vehicles that made contact while following behind a Waymo ADS vehicle. The Waymo vehicle was not contacted or damaged. Second, case 30270-6341 involved a Waymo ADS vehicle stopped near the entrance of an intersection yielding to a fire engine. While the fire engine was navigating around the stopped Waymo vehicle, the fire engine allegedly contacted another vehicle also stopped in a through lane. Because the purpose of this paper is to compare crashed vehicle rates between ADS and human benchmarks, we excluded these crashes from the NHTSA SGO in-transport vehicle category and subsequent categories. These two events were included in the SGO-reported count.

Police-reported status was determined using a combination of the NHTSA SGO field ``Law Enforcement Investigating?'' and internal Waymo records. There were 10 SGO-reported crashes that indicated ``Yes'' for the ``Law Enforcement Investigating?'' field. The two (2) SGO-reported crashes with an ``Unknown'' reported in the ``Law Enforcement Investigating?'' field also likely had a police report generated based on review of Waymo operations records. Additionally, in two events (30270-6548 and 30270-1583), the SGO field ``Law Enforcement Investigating?'' is ``No'', but operational records indicate a police report may have been filed. These two events were also conservatively included in the SGO police-reported category. Event \#1 from \citet{victor2023_1mRO} occurred before the NHTSA SGO-reporting period but was reported to police based on review of operational records. To determine \textit{any-injury-reported} crashes, the ``Highest Injury Severity Alleged'' SGO field was used. Any crash with ``Minor'', ``Moderate'', ``Serious'', or ``Fatality'' as the maximum severity was considered an \textit{any-injury-reported} crash. In this dataset, there were 3 crashes with ``Minor'' injuries reported. In addition, there was one crash (30270-6149) with ``Unknown'' maximum severity. The written SGO case narrative for this case states there were reported injuries and a person involved in the crash ``was transported from the scene to a hospital for medical treatment.'' This crash with ``Unknown'' severity was considered an \textit{any-injury-reported} crash in this study, bringing to total injury crash count to 4. Other crashes with ``Unknown'' maximum severity due to missing information (e.g., due to another party leaving the scene without exchanging information) were excluded from the \textit{any-injury-reported} outcome, as these types of missing data are also excluded from the human crash databases used to derived the benchmarks used in this study \citep{scanlon2023benchmark}. Crashes where the ADS was not-in-transport were also excluded from \textit{police-reported} and alleged \textit{any-injury-reported} crashes. All ADS not-in-transport crashes had no reported law enforcement investigation and no injuries reported as part of the SGO data.

This study relied on fields in the NHTSA SGO crash data to determine ADS outcome groups. There may be differences in how different ADS fleet operators fill out the fields in the NHTSA SGO reports. For example, other ADS fleet operators may interpret the ``Law Enforcement Investigating?'' field differently than what was described in this study.

\begin{longtable}[c]{p{0.8in}p{0.8in}p{0.5in}p{0.5in}p{0.5in}p{0.5in}p{0.5in}p{0.5in}}
  \toprule
  \multicolumn{8}{c}{\textbf{List of Included NHTSA SGO Crash Events in this Analysis.}} \\
  \textbf{SGO Report ID} & \textbf{Waymo 1M Mile RO Event \# \citep{victor2023_1mRO}} & \textbf{Location} & \textbf{NHTSA SGO-Reported} & \textbf{NHTSA SGO In-Transport} & \textbf{NHTSA SGO In-Transport Exclude Low DV} & \textbf{NHTSA SGO Police-Reported} & \textbf{NHTSA SGO Any Injury} \\
  \midrule
  \endfirsthead
  
  \toprule
  \multicolumn{8}{c}{\textbf{Continuation of Event List}} \\
  \textbf{SGO Report ID} & \textbf{Waymo 1M Mile RO Event \# \citep{victor2023_1mRO}} & \textbf{Location} & \textbf{NHTSA SGO-Reported} & \textbf{NHTSA SGO In-Transport} & \textbf{NHTSA SGO In-Transport Exclude Low DV} & \textbf{NHTSA SGO Police-Reported} & \textbf{NHTSA SGO Any Injury} \\
  \midrule
  \endhead
 
  \bottomrule 
  \endfoot
 
  \bottomrule
  \endlastfoot
 
  30270-6668 & NA & PHX & True & True & True & False & False \\
  30270-6667 & NA & SFO & True & True & False & False & False \\
  30270-6666 & NA & SFO & True & True & False & False & False \\
  30270-6664 & NA & PHX & True & True & False & False & False \\
  30270-6663 & NA & SFO & True & True & True & False & False \\
  30270-6662 & NA & SFO & True & True & False & False & False \\
  30270-6661 & NA & SFO & True & True & True & False & False \\
  30270-6566 & NA & PHX & True & True & False & True & True \\
  30270-6561 & NA & PHX & True & True & False & True & False \\
  30270-6548 & NA & PHX & True & True & False & True & False \\
  30270-6542 & NA & PHX & True & False & False & False & False \\
  30270-6521 & NA & PHX & True & True & False & False & False \\
  30270-6520 & NA & PHX & True & True & False & False & False \\
  30270-6519 & NA & PHX & True & True & True & True & False \\
  30270-6518 & NA & SFO & True & False & False & False & False \\
  30270-6517 & NA & SFO & True & True & False & False & False \\
  30270-6516 & NA & PHX & True & True & True & False & False \\
  30270-6515 & NA & SFO & True & True & True & False & False \\
  30270-6514 & NA & SFO & True & True & False & False & False \\
  30270-6405 & NA & SFO & True & True & True & True & False \\
  30270-6385 & NA & SFO & True & True & True & False & False \\
  30270-6368 & NA & PHX & True & True & True & True & False \\
  30270-6352 & NA & SFO & True & True & True & True & False \\
  30270-6347 & NA & LA & True & True & True & False & False \\
  30270-6346 & NA & SFO & True & True & False & False & False \\
  30270-6342 & NA & SFO & True & True & False & False & False \\
  30270-6341 &   & SFO & True & False & False & False & False \\
  30270-6338 & NA & PHX & True & True & True & False & False \\
  30270-6337 & NA & PHX & True & False & False & False & False \\
  30270-6336 & NA & SFO & True & True & True & True & True \\
  30270-6335 & NA & PHX & True & False & False & False & False \\
  30270-6198 & NA & PHX & True & True & True & True & False \\
  30270-6150 & NA & PHX & True & True & True & False & False \\
  30270-6149 & NA & PHX & True & True & True & True & True \\
  30270-6133 & NA & SFO & True & True & False & False & False \\
  30270-6132 & NA & SFO & True & True & False & False & False \\
  30270-6131 & NA & PHX & True & False & False & False & False \\
  30270-5997 & NA & PHX & True & True & True & True & True \\
  30270-5961 & NA & SFO & True & True & False & False & False \\
  30270-5960 & NA & SFO & True & False & False & False & False \\
  30270-5959 & NA & SFO & True & True & True & False & False \\
  30270-5958 & NA & SFO & True & False & False & False & False \\
  30270-5957 & NA & SFO & True & True & False & False & False \\
  30270-5896 & NA & PHX & True & True & True & True & False \\
  30270-5760 & NA & PHX & True & True & False & False & False \\
  30270-5758 & NA & SFO & True & True & True & False & False \\
  30270-5756 & NA & PHX & True & True & False & False & False \\
  30270-5612 & NA & PHX & True & True & False & False & False \\
  30270-5611 & NA & PHX & True & True & False & False & False \\
  30270-5593 & NA & SFO & True & False & False & False & False \\
  30270-5592 & NA & SFO & True & True & True & False & False \\
  30270-5591 & NA & PHX & True & True & True & False & False \\
  30270-5588 & NA & PHX & True & True & False & False & False \\
  30270-5456 & NA & SFO & True & True & True & False & False \\
  30270-5319 & NA & PHX & True & True & False & False & False \\
  30270-5318 & NA & SFO & True & True & False & False & False \\
  30270-5208 & NA & SFO & True & True & False & False & False \\
  30270-5114 & NA & SFO & True & True & False & False & False \\
  30270-5085 & NA & SFO & True & True & True & False & False \\
  30270-5083 & NA & SFO & True & True & True & False & False \\
  30270-5081 & NA & SFO & True & True & False & False & False \\
  30270-4882 & NA & PHX & True & True & True & False & False \\
  30270-4880 & 19 & PHX & True & True & False & False & False \\
  30270-4768 & 20 & PHX & True & True & True & False & False \\
  30270-4484 & 16 & PHX & True & True & False & True & False \\
  30270-4363 & 17 & PHX & True & True & False & False & False \\
  30270-3842 & 14 & SFO & True & True & True & False & False \\
  30270-3838 & 13 & PHX & True & True & False & False & False \\
  30270-1730 & 6 & PHX & True & False & False & False & False \\
  30270-1583 & 7 & PHX & True & True & True & True & False \\
  30270-1501 & 5 & PHX & True & True & True & False & False \\
  NA & 4 & PHX & True & True & True & False & False \\
  NA & 1 & PHX & True & True & True & True & False \\
  \end{longtable}

As stated in the Methods section, the ``NHTSA SGO In-Transport Exclude Low delta-V (DV)'' category was derived by applying a impulse-momentum collision model for vehicle-to-vehicle crashes. The collision model used was an impulse-momentum model using the inertial properties of the ADS equipped vehicle and estimated inertial properties of the crash partner, as described in \citet{scanlon2021fatal}. Each of the 7 crashes with fixed or non-fixed objects was examined individually to estimate a delta-V. Of the 7 crashes with fixed or non-fixed objects, 5 were excluded for having a low delta-V. Crashes where the ADS was not-in-transport were also excluded from this group. Three (3) fixed object crashes resulted in underbody damage when the Waymo vehicle ``bottomed out'' entering a parking lot driveway (30270-5612, 30270-5611) or tire damage after driving through a pothole (30270-5114) which resulted in less than the 1.0 mph delta-V threshold. Two (2) crashes involved non-fixed objects with low mass (<100 kg): a cardboard box (30270-5081) and a swinging parking lot gate that was not fixed (30270-4363).  The single crash involving a cyclist (30270-5456) was included in this group. 

Of the two vehicle-to-object crashes included, SGO report 30270-6548 involved a Waymo ADS vehicle that was driving in a construction zone and ``entered a lane undergoing construction ..., encountered a section of roadway that had been removed, and the front driver’s side wheel dropped off the paved roadway.'' After reviewing the on-board data from this crash event, the front wheel dropping likely caused a delta-V of 1 to 3 mph. SGO report 30270-6561 involved a Waymo ADS vehicle ``exited a parking lot..., when the passenger’s side sensors and side mirror made contact with an automatic gate that was in the process of closing.'' Based on review of the sensor data, the contact with the closing gate caused damage to a sensor housing and side view mirror of the vehicle, which are unlikely to result in a delta-V above 1 mph. Because the event in report 30270-6561 was police-reported, we decided to include this event from the low delta-V category.

All of the crash rate benchmarks presented in this study are at the vehicle-level (or driver-level crash rates), and are described as a crashed vehicle rate. Crashed vehicle rates represent the rate at which drivers (or a driver) crash(es) per VMT. The physical representation of a vehicle-level crash rate is equivalent to how ADS crash rates are presented - the rate at which the ADS crashes per VMT. It is notable that a common mathematical error is pervasive in the ADS benchmark literature that incorrectly compares a crash-level crash rate from the human benchmarks to a vehicle-level crash rate from the ADS data \citep{scanlon2023benchmark}. As a simple illustration, consider two drivers (driver 1 and 2) that each travels 100 miles and collides. The crash rate in this population is 0.5 crashes per 100 miles (1 crash / 200 miles) and the crashed vehicle rate is 1 crashed vehicle per 100 miles (2 crashed vehicles / 200 miles traveled). Now consider a hypothetical ADS (driver 3) that travels 100 miles and crashes into another vehicle (driver 4), for which the miles traveled is known. This crashed ADS vehicle rate of 1 crashed vehicle per 100 miles is comparable to the benchmark crashed vehicle rate of driver 1 and 2 of 1 crashed vehicle per 100 miles. By contrast, comparing the ADS crashed vehicle rate of 1 crashed vehicle per 100 miles to the human-driven crash rate of 0.5 crashes per 100 miles would produce a faulty comparison that ignores the number of vehicles involved in the calculation of the crash rate.

\subsection{Results for Los Angeles}\label{apdx:la_results}
In Los Angeles, there was a single SGO-Reported crash during the study period. This single event also met the in-transport and exclude low delta-V inclusion criteria resulting in an IPMM of 21.4 with 95\% confidence intervals of (0.5, 119) IPMM. There were no (zero) \textit{police-reported} and \textit{any-injury-reported} crashes in Los Angeles, resulting in a 0 IPMM with 95\% confidence intervals of (0, 64.1) IPMM.

The 1 crash in Los Angeles met both the SGO and low delta-V thresholds in the \textit{any property damage or injury} comparison. Compared to the Blincoe-adjusted benchmark, the rate ratio was 2.84 with confidence intervals of (0.04, 18.1). Compared to the Blanco-adjusted benchmark, the rate ratio was 1.40 with confidence intervals of (0.02, 8.92). Due to the low ADS miles driven, neither comparison was statistically significant.

There were no observed \textit{police-reported} or \textit{any-injury-reported} crashes in Los Angeles resulting in an rate ratio of 0. The result in Los Angeles was not statistically significant due to low mileage (confidence intervals of [0, 21.3] for \textit{police-reported} and [0, 32.8] for \textit{any-injury-reported}). 

\subsection{Example of \citet{nelson1970confidence} Rate Ratio Confidence Intervals}\label{apdx:nelson_cis}

An example of the \citet{nelson1970confidence} confidence interval for the crash rate ratio is shown in Listing~\ref{lst:nelson_function} in the Python language.

\begin{lstlisting}[caption=Python Function to Compute \citet{nelson1970confidence} Rate Ratio Confidence Intervals, label={lst:nelson_function}, language=Python]
from scipy.stats import betaprime

def nelson_ci(X, s, Y, t, alpha=0.05):
  """Clopper-Pearson limits for a rate ratio.

  Args:
    X: Benchmark event count
    s: Benchmark exposure (e.g., miles)
    Y: ADS Event Count
    t: ADS exposure (e.g., miles, matching benchmark exposure)
    alpha: alpha level

  Returns:
    Rate Ratio and (lower, upper) confidence intervals
  """
  if Y > 1e-3:
    # 2-sided test, except for zero events (then it's one sided)
    alpha = alpha / 2

  point =  (Y / t) / (X / s) if X > 0 else float("inf")
  lower = s / t * betaprime.ppf(alpha / 2, Y, X + 1) if Y > 0 else 0
  upper = s / t * betaprime.ppf(1.0 - alpha / 2, Y + 1, X)

  return point, (lower, upper)
\end{lstlisting}

Listing~\ref{lst:nelson_tests} shows test cases based on select results from Table~7. Listing~\ref{lst:nelson_tests_output} shows the output of the test code.

\begin{lstlisting}[caption=Test Cases to Compute \citet{nelson1970confidence} for Example Results from Table~7, label={lst:nelson_tests}, language=Python]
import sys
import scipy
print(f"Python version: {sys.version_info}")
print(f"scipy version:{scipy.__version__}\n")

# (Label, Benchmark IPMM, Benchmark Million Miles, Waymo Count,
# Waymo Million Miles) from Table 2 (benchmark) and Table 4 (Waymo)
phx_benchmark_m_miles = 24865
sf_benchmark_m_miles = 862
phx_waymo_m_miles = 5.343
sf_waymo_m_miles = 1.755
test_cases = [
    ("Police-reported PHX", 4.31, phx_benchmark_m_miles, 12, phx_waymo_m_miles),
    ("Police-reported SF", 5.86, sf_benchmark_m_miles, 3, sf_waymo_m_miles),
    ("Any-injury PHX", 1.81, phx_benchmark_m_miles, 3, phx_waymo_m_miles),
    ("Any-injury SF", 5.82, sf_benchmark_m_miles, 1, sf_waymo_m_miles),
]

# (Rate Ratio, CI Lower, CI Upper) as reported in the Table 7
expected = [
    (0.52, 0.24, 0.97),
    (0.29, 0.05, 0.95),
    (0.31, 0.05, 1.01),
    (0.10, 0.001, 0.62),
]

# results are round to 2 decimal places
test_tol = 1e-2
for test_input, test_expected in zip(test_cases, expected):
  (label, benchmark_ipmm, benchmark_m_miles, waymo_count, waymo_m_miles) = (
      test_input)
  (expected_point, expected_ci_lower, expected_ci_upper) = test_expected
  benchmark_count = benchmark_ipmm * benchmark_m_miles
  rate_ratio, (ci_lower, ci_upper) = nelson_ci(
      benchmark_count, benchmark_m_miles, waymo_count, waymo_m_miles)
  print(f"Comparison: {label}, Rate Ratio: {rate_ratio:.2f}, "
        f"CI lower: {ci_lower:.2f}, CI upper: {ci_upper:.2f}")
  assert abs(rate_ratio - expected_point) < test_tol
  assert abs(ci_lower - expected_ci_lower) < test_tol
  assert abs(ci_upper - expected_ci_upper) < test_tol
\end{lstlisting}

\begin{lstlisting}[caption=Output of Test Cases, label={lst:nelson_tests_output}, language=Python]
Python version: sys.version_info(major=3, minor=11, micro=8, releaselevel='final', serial=0)
scipy version:1.13.1

Comparison: Police-reported PHX, Rate Ratio: 0.52, CI lower: 0.24, CI upper: 0.97
Comparison: Police-reported SF, Rate Ratio: 0.29, CI lower: 0.05, CI upper: 0.95
Comparison: Any-injury PHX, Rate Ratio: 0.31, CI lower: 0.05, CI upper: 1.01
Comparison: Any-injury SF, Rate Ratio: 0.10, CI lower: 0.00, CI upper: 0.63
\end{lstlisting}

\subsection{Considerations for Safety Impact Assessment for Serious Injury and Fatal Crashes}\label{apdx:serious_and_fatal}

Conclusions regarding serious injury alone, or fatalities alone, require more data because they are subsets of the \textit{any-injury-reported} outcome group which combines all injury and fatality results into one data category. ADS developers use multiple, complementary safety assurance methods to mitigate risk of high severity outcomes across architectural, behavioral, and operational layers (e.g., platform verification and validation, hazard analysis, see \citealt{webb2020readiness}, \citealt{favaro2023building}). Traditionally, automotive safety systems have used prospective safety benefit analyses that use test track data and simulations as risk assessment tools and to determine potential high severity benefits. These types of prospective studies have also been performed for ADS. For example, one study that used reconstructions of all fatal crashes in Chandler, AZ found that a level 4 ADS was able to prevent 100\% of crash when placed in the initiator role and prevent 82\% and mitigate 10\% when placed in the responder role \citep{scanlon2021fatal}. 

Furthermore, ADS are designed to be able to follow applicable speed limits on the roads they drive on. Reducing speeding is a core pillar of the Vision Zero and Safe Systems approach to reducing serious injuries and fatalities on roadways. The Safe Systems approach has been adopted as a policy initiative by the U.S. Department of Transportation \citep{usdot2022nrss}. The results of this study show that the ADS evaluated had a statistically significant lower \textit{any-injury-reported} crashed vehicle rate. Serious injury and fatalities are a subset of this \textit{any-injury-reported} benchmark, but no statement on these outcome levels can be made at this time based on this retrospective data. In a Safe Systems approach, it is not a requirement to prove efficacy of countermeasures before they are deployed in the field. Due to the statistical requirements, like the ones presented in \citet{scanlon2023benchmark} or \citet{kalra2016driving}, assessing the safety impact of a system on serious injury or fatal crashes is impractical in many situations. The Safe Systems approach instead uses known energy-related factors that contribute to a large amount of serious injury and fatal crashes (e.g., excessive speed, or excessive energy transfer) and attempts to control these factors to stay below the tolerances to lead to these outcomes. 

As ADS RO miles increase, the likelihood of observing a serious injury or fatality also increases. As discussed in the statistical power analysis of \citet{scanlon2023benchmark}, hundreds of millions to billions of miles of VMT would be necessary to detect a statistically significant difference in fatal crashed vehicle rates, depending on the magnitude of the difference in crashed vehicle rate of the ADS and benchmark. The national rate of crashed passenger vehicles involved in fatal crashes while traveling on surface streets is 22.3 incidents per billion miles (IPBM) \citep{scanlon2023benchmark}. Consider a fictive driver who experiences 1 fatal crash after driving 10 million miles. At the time of this fictive fatal crash, the driver’s fatal crashed vehicle rate would be 100 IPBM with 95\% confidence intervals of [3, 557] IPBM. Although the point estimate of 100 IPBM is higher than the benchmark, due to the low event count relative to the miles, this fictive driver’s crashed vehicle rate is not statistically different from the benchmark. Consider another fictive driver involved in 1 fatal crash in 100 million VMT, which is a rate of 10 IPBM. The 95\% confidence intervals of this fictive driver is [0.3, 56] IPBM, which is still overlapping with the benchmark and not significantly different.

\end{document}